\documentclass[10pt,journal,compsoc]{IEEEtran}

\ifCLASSOPTIONcompsoc
  \usepackage[nocompress]{cite}
\else
  \usepackage{cite}
\fi
\ifCLASSINFOpdf
\else
\fi
\usepackage{amsmath}
\usepackage{subfig}
\usepackage{graphicx}
 \usepackage{multirow}
 \usepackage[table,xcdraw]{xcolor}
 \usepackage{amsmath,amssymb}

\usepackage{algpseudocode}
\usepackage{algorithm}

\algnewcommand\algorithmicforeach{\textbf{for each}}
\algdef{S}[FOR]{ForEach}[1]{\algorithmicforeach\ #1\ \algorithmicdo}

\begin{document}
\title{Exploiting Human Social Cognition for the Detection of Fake and Fraudulent Faces via Memory Networks}

\author{Tharindu~Fernando,~\IEEEmembership{Student Member,~IEEE,}
	Clinton~Fookes,~\IEEEmembership{Senior Member,~IEEE,}
	Simon~Denman,~\IEEEmembership{Member,~IEEE,}
	~and~ Sridha~Sridharan,~\IEEEmembership{Life Senior Member,~IEEE.}
        \IEEEcompsocitemizethanks{\IEEEcompsocthanksitem T. Fernando, C. Fookes, S.Denman and S. Sridharan   are with Image and Video Research Lab, SAIVT, Queensland University of Technology, Australia.\protect\\
        }}

%
%

\markboth{Journal of \LaTeX\ Class Files,~Vol.~14, No.~8, August~2015}%
{Fernando \MakeLowercase{\textit{et al.}}: Bare Demo of IEEEtran.cls for Computer Society Journals}
%



\IEEEtitleabstractindextext{%
\begin{abstract}
Advances in computer vision have brought us to the point where we have the ability to synthesise realistic fake content. Such approaches are seen as a source of disinformation and mistrust, and pose serious concerns to governments around the world. Convolutional Neural Networks (CNNs) demonstrate encouraging results when detecting fake images that arise from the specific type of manipulation they are trained on. However, this success has not transitioned to unseen manipulation types, resulting in a significant gap in the line-of-defense. We propose a Hierarchical Memory Network (HMN) architecture, which is able to successfully detect faked faces by utilising  knowledge stored in neural memories as well as visual cues to reason about the perceived face and anticipate its future semantic embeddings. This renders a generalisable face tampering detection framework. Experimental results demonstrate the proposed approach achieves superior performance for fake and fraudulent face detection compared to the state-of-the-art.
\end{abstract}

\begin{IEEEkeywords}
Neural Memory Networks, Fake and Fraudulent Face Detection, Image and Video Forensics.
\end{IEEEkeywords}}

\maketitle

\IEEEdisplaynontitleabstractindextext

%
\IEEEpeerreviewmaketitle

\IEEEraisesectionheading{\section{Introduction}\label{sec:introduction}}

\IEEEPARstart{W}{ith} social media becoming a primary source of information for many people, and with more than 100 million hours of video content watched daily \cite{afchar2018mesonet}, fake news has quickly risen as a significant threat to society and democracy. As demonstrated by the DeepFake app \cite{deepfake_app}, the effortless and seamless nature of synthesising realistic fake content could have a crucial impact on people's lives and severely undermine the trust in digital media \cite{rossler2018faceforensics,deepfakes_impact}. In addition, for different kinds of applications such as biometric-based authentication \cite{jain2004introduction,ferrara2014magic} filtering out forged images and videos is a prominent issue \cite{rossler2018faceforensics}. 

While deep neural networks \cite{bayar2016deep,rahmouni2017distinguishing,rossler2018faceforensics,afchar2018mesonet} have attained striking levels of accuracy when detecting certain types of image tampering attacks, their performance is often disappointing when presented with images or videos with unseen manipulation methods \cite{nguyen2018capsule,cozzolino2018forensictransfer}. It is observed that the underlying neural network can quickly overfit to a specific artefact left by the tampering method observed during training \cite{cozzolino2018forensictransfer}, and thus methods lack transferability. Furthermore, methods for image tampering detection are not directly transferable to videos due to artefacts and quality loss resulting from the video compression. As such, there exists a critical need for automated systems which capture the fundamentals of video and image content tampering, and have the ability to generalise to different manipulation attacks.

In this paper, we propose a deep learning framework to detect fake and fraudulent faces in images and video which is motivated by the social perception \cite{allison2000social} and social cognition \cite{fiske2013social} processes in the human brain. Recent neuroscientific research \cite{schindler2017differential,wheatley2011mind} has reported that humans try to predict the observed person's mental state through visual cues such as their facial expression and by utilising specific knowledge stored in the observer's memories. Furthermore, the authors of \cite{wheatley2011mind} conclude that during this prediction process, the lack of social presence and social attributes cause the uncanny valley effect \cite{schindler2017differential} in the brain when humans view computer generated and tampered media. Inspired by the uncanny valley effect of the human brain, we propose a deep learning framework for detecting fake and fraudulent faces where we predict the semantic embeddings of the future face state, enforcing the model to learn the social context of the perceived face. However, we also know the human brain is far from infallible for such a task as it does fail to detect cases of sophisticated tampering and synthetic face generation. To combat this, we also leverage the success of multi-task learning to jointly predict if a face is real or fake, along with the prediction of the future face state. The final detector, while inspired by human cognition, does not directly replicate the process of humans. This results in a highly effective approach to detect face and fraudulent faces which is also generalisable to unseen manipulation types. 
 
 We exploit Neural Memory Networks (NMNs) to facilitate the above two tasks by mapping long-term dependencies in the data domain. Current state-of-the-art NMNs such as the Neural Turing Machine (NTM) \cite{graves2014neural} and Tree Memory Networks (TMNs) \cite{fernando2018tree} have demonstrated encouraging results when the models are required to maintain additional capacity to recover long-term dependencies. However, through proper comparative illustrations and evaluations we clearly demonstrate that the current state-of-the-arts do not optimally map these long-term dependencies due to architectural deficiencies and propose a novel Hierarchical Memory Network (HMN) architecture to overcome these limitations.  
 
In addition, as there doesn't exist an optimal, off the shelf loss function for joint learning of the fake face detection and future semantic embedding anticipation tasks, we propose an adversarial learning platform to automatically learn a loss function that considers both tasks. 
 
\begin{figure}[htbp]
\centering
 \includegraphics[width=1\linewidth]{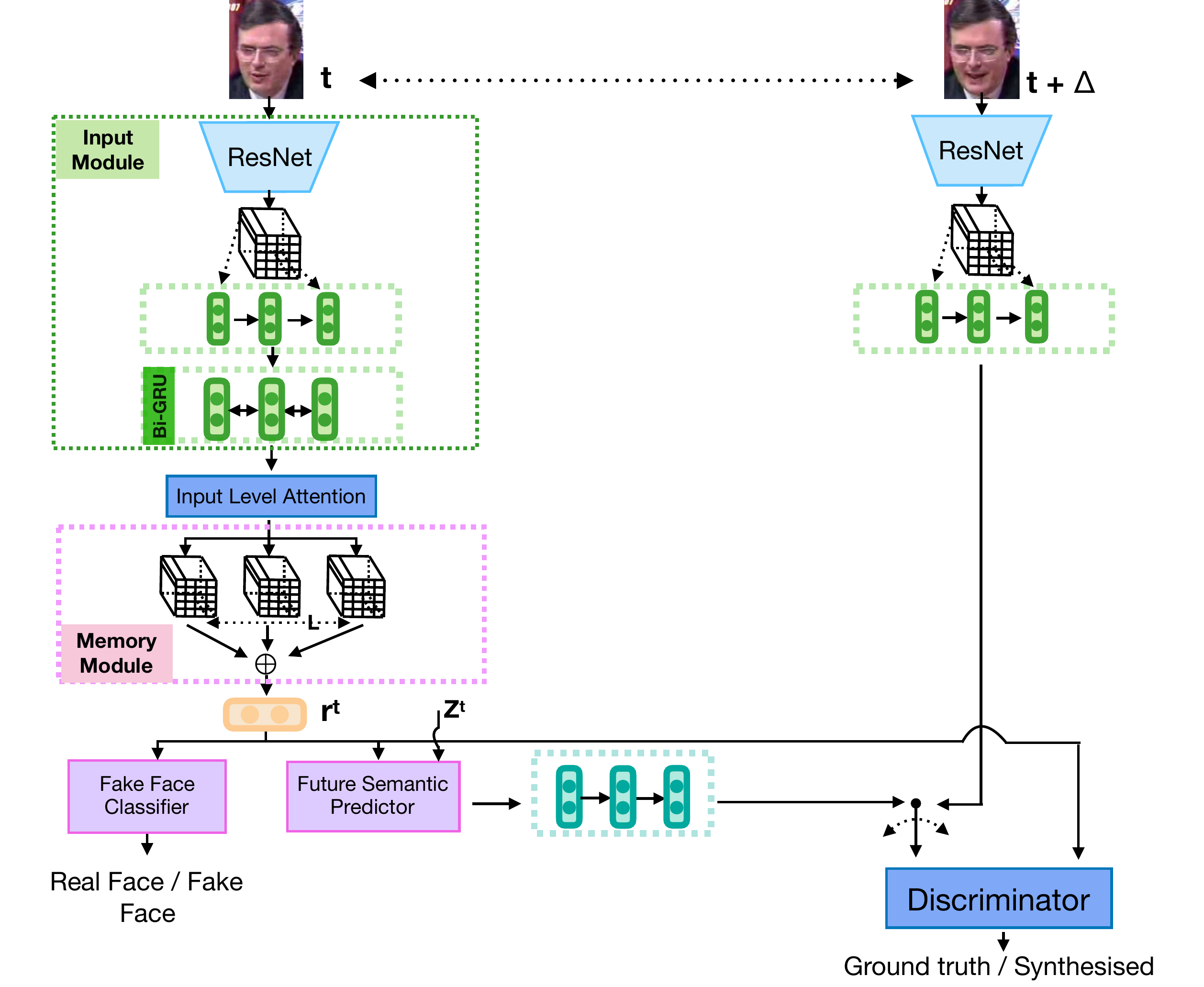}
\caption{Proposed Hierarchical Memory Network (HMN) framework for tampered face detection and future face semantic embedding prediction: In the \textbf{input module} visual facial features are extracted through a pre-trained ResNet \cite{he2016deep}. The extracted embeddings are rearranged in a sequence and passed it through a bi-directional GRU \cite{cho2014learning} to map their relationships. \textbf{Input level attention} is then employed to extract informative embeddings, deriving a vector to query the memory. The \textbf{memory module} outputs information regarding the authenticity of the face region and it's future behaviour. In the learning process we utilise a discriminator network which either receives ground truth CNN semantic embeddings $\Delta$ frames ahead, or synthesised semantic embeddings from the memory output, from which it learns to classify the embeddings are real or fake. The future semantic predictor tries to fool the discriminator in the process, and this adversarial loss is coupled with a fake face classification loss to form the complete objective of the HMN.}
\label{fig:overview}
\end{figure}

An overview of the proposed framework is presented in Fig. \ref{fig:overview}. Given an input image of a face, we first extract an embedding that captures facial semantics using a ResNet \cite{he2016deep} model pre-trained on ImageNet \cite{deng2009imagenet}. This semantic embedding is used to query the memory. The memory output is used for two tasks: (i) for classification of the authenticity of the face; and (ii) the prediction of a future face semantic embedding. Facilitating the adversarial learning framework, the predicted semantic embeddings are passed through a discriminator. During offline training, the discriminator receives either a predicted or a ground truth future embedding, alongside the same input frame as the generator, and learns to classify real embeddings from fake. This training process requires pairs of face images and their representation at future time-steps. For testing, it can be applied to authenticate either images or videos.  
 
 To the best of our knowledge, this is the first work employing NMNs in the context of multimedia forensics as well as the first work attempting multi-task learning of NMNs through adversarial training.  To this end we make the following contributions: 
 \begin{itemize}
 \item We propose a method to detect tampered face images that is inspired by the human social cognition process, and uses joint learning to predict the future face state, supporting the primary tampered media detection task.
 \item We propose a novel neural memory architecture for hierarchal semantic embedding propagation and prediction.
 \item We introduce an adversarial learning paradigm to train the memory architecture where the model learns a task specific loss for both fake face classification and future semantic generation tasks.
 \item We perform extensive evaluations on the proposed network using multiple public benchmarks where we outperform baseline methods by a substantial margin, especially for unseen tampering categories.
 \item We analyse what is being activated in the proposed hierarchical memory models during learning and interpret the learning process.    
 \end{itemize}

Considering the critical need for a fully automated system for the detection of fake, fraudulent, and tampered faces, we demonstrate the proposed system on its primary application of face tampering detection. However, its applications are not constrained to this domain and the framework is directly applicable to any machine learning setting which requires the hierarchical capture of long-term dependencies. 

\section{Related Work}
Related work within the scope of this paper can be categorised into human social perception and cognition (Sec. \ref{sec:lit_human}),  face manipulation techniques (Sec. \ref{sec:lit_face_manu}), face manipulation detection techniques (Sec. \ref{sec:lit_face_manu_det}), and memory architectures (Sec. \ref{sec:lit_mem}). 

\subsection{Human Social Perception and Cognition}
\label{sec:lit_human}
Recent neuroscientific research \cite{schindler2017differential,wheatley2011mind} has investigated human brain activities related to the uncanny valley \cite{schindler2017differential} effect, where humans report an ``eerie feeling'' when observing faces which lack natural attributes like emotions. Their observations suggest a two-stage process in human face cognition. In the first phase, the visual input is matched to a face template which is denoted by strong activities in dedicated areas of the visual cortex (i.e occipital face area \cite{pitcher2007tms} and fusiform face area \cite{kanwisher1997fusiform}). In the second phase, they detected a subsequent activation as the observed face is evaluated for social perception. Most interestingly, they conclude that the observed face region is evaluated by inferring the state of the observed person's mind \cite{schindler2017differential}. Through visual cues such as facial expression and by utilising specific knowledge stored in observer's memories, humans try to understand the observed person's mental state (e.g. he or she is is angry). Subsequently, through a human's natural social cognition processes, they build a theory about that particular person and their emotions (e.g. why is he or she angry?). Most importantly these processes are applied even if the observed face is not familiar to the perceiver. In such cases, the inferences are indexed based on visual cues in their semantic memory for later use \cite{wheatley2011mind}. In addition, \cite{wheatley2011mind} illustrates that computer generated and tampered media generates the uncanny valley effect due to the lack of social presence and social attributes. In the proposed work we exploit this aspects of human social perception and cognition process. However, instead of directly replicating the two phase process of humans, we utilise the recent advances in machine learning to automatically detect fake and fraudulent faces, which are designed to fool the humans.

\subsection{Face Manipulation Techniques}
\label{sec:lit_face_manu}

From one of the first attempts reported in Dale et. al \cite{dale2011video} video face replacement has rapidly advanced. In \cite{garrido2014automatic} the authors investigated methods to preserve expressions while replacing faces. The method of \cite{garrido2015vdub} uses 3D face capture to match the mouth movements between the actors and the dubber. 

One of the most notable methods among video face replacement techniques is DeepFake \cite{deepfake_app} which aims to replace a face in an input image with a target face. The core idea behind the DeepFake generation is to train two auto-encoders, with a common encoder, in parallel \cite{afchar2018mesonet}. The goal of the auto-encoders is to represent the input image with a reduced number of parameters and output an approximation of the original input. 


At test time using the shared encoder and the decoder of $B$ one can generate a mapping from input $A$ to a synthesised image $B$. During encoding, the encoder is expected to segregate fundamental facial attributes from other information such as the background. Hence during decoding, the decoder should map these facts to the target face considering its contextual information. 

The Face2Face \cite{thies2016face2face} method transfers facial expressions from an input to a target frame. The technique first reconstructs and tracks the input source and target faces through a dense tracking system. Then using the expression translation mechanism proposed in \cite{thies2016face2face} the authors translate the expressions of the source face to the target face. 

In a different line of work the authors in \cite{kim2018deep} propose an image-to-image translation where they learned a mapping between the faces from computer graphics and real face images. Suwaganakorn et. al \cite{suwajanakorn2017synthesizing} propose a mapping between audio and lip movements and Averbuch-Elor et. al \cite{averbuch2017bringing} propose a system to bring portraits to life. 

Considering the recent success of GANs for synthesising realistic content, numerous such methods have been proposed to alter face images. For instance, in \cite{lample2017fader,upchurch2017deep} authors propose to alter face attributes including age, moustaches and smiling. A method to improve the image quality of the synthesised content is proposed in \cite{karras2017progressive}. Most recently, Karras et. al \cite{karras2019style} propose a GAN based face generator which learns to alter high level facial attributes such as pose and identity, as well as the stochastic variation of features such as freckles and hair in the synthesised content. 

These techniques have demonstrated their superior ability to generate realistic-looking faces. However, the authors in \cite{afchar2018mesonet} showed that the synthesised faces from DeepFake lack the expressiveness of real examples. Furthermore, considering the results presented in \cite{thies2016face2face,rossler2018faceforensics}, we observe that the process introduces unnatural attributes to different face regions. Hence, the proposed future face semantic prediction based framework facilitates an ideal platform to learn a conditional distribution to segregate examples that contain those manipulations from those that do not.

\subsection{Face Manipulation Detection Techniques}
\label{sec:lit_face_manu_det}

There exist several CNN based methods to detect image manipulations. For instance, in \cite{bayar2016deep} the authors introduce a novel predictive kernel to detect pixels that deviate from the logical local structural relationships in the neighbourhood. The kernel first predicts the pixel value of the centre pixel in the given window and subtracts the actual pixel value to get the prediction error. These prediction errors are propagated to subsequent layers. Quan et. al \cite{quan2018distinguishing} proposed the use of 3D convolutional kernels to learn discriminative features and separate natural images from computer generated graphics. In a different line of work Li et. al \cite{li2018ictu} proposed a method to detect fake video content by solely utilising eye blink patterns. They show that real video content contains a unique eye blinking pattern which fake content lacks. In \cite{nguyen2018capsule} the authors investigate the efficiency of capsule networks for fake content detection. 

However, as shown in \cite{cozzolino2018forensictransfer}, CNN based methods tend to overfit to a particular type of artefact, limiting the applicability of these methods for general biometric and security applications. Furthermore, reenactment techniques do not carry the artefacts of blinking patterns as they generally tamper with the nose and mouth regions. Hence, the general applicability of \cite{li2018ictu} is questionable. The authors in \cite{sabir2019recurrent} exploit both spatial and temporal information through utilising a recurrent convolutional model. 

A method that considers transferability in face forgery detection is presented in \cite{cozzolino2018forensictransfer}. They consider an encoder-decoder framework and explicitly force the encoder to learn two separate encoding spaces for real and fake images. As a result, the total loss of their system is the combination of this classification loss and the reconstruction loss of the decoder. 

In a different line of work researchers and have investigated the possibility of utilising inconsistency between the claimed camera parameters and the characteristics of perspective distortion \cite{wang2017position} and illumination conditions \cite{peng2016optimized} to determine the integrity of input images. A multi-task learning paradigm is proposed in \cite{nguyen2019multi} where the authors combine the fake face classification task together with the segmentation of manipulated regions. The authors designed the network to share information between the tasks, however they did not observe a significant performance improvement from the joint learning of both tasks. 

\subsection{Memory Architectures}
\label{sec:lit_mem}
With the tremendous success achieved by recurrent neural networks such as Long Short-term Memory (LSTM) \cite{hochreiter1997long} networks and Gated Recurrent Units (GRUs) \cite{cho2014learning}, numerous works have employed what are termed ``memory modules''. Memory modules are expected to store important facts and map long-term dependencies when questioned with a particular input. However, experiments conducted in \cite{fernando2018soft+,fernando2018tree} demonstrated that the memory modules in LSTMs and GRUs only map relationships within an input sequence, and disregard dependencies between different inputs. 

\begin{figure}[htbp]
\centering
 \includegraphics[width=\linewidth]{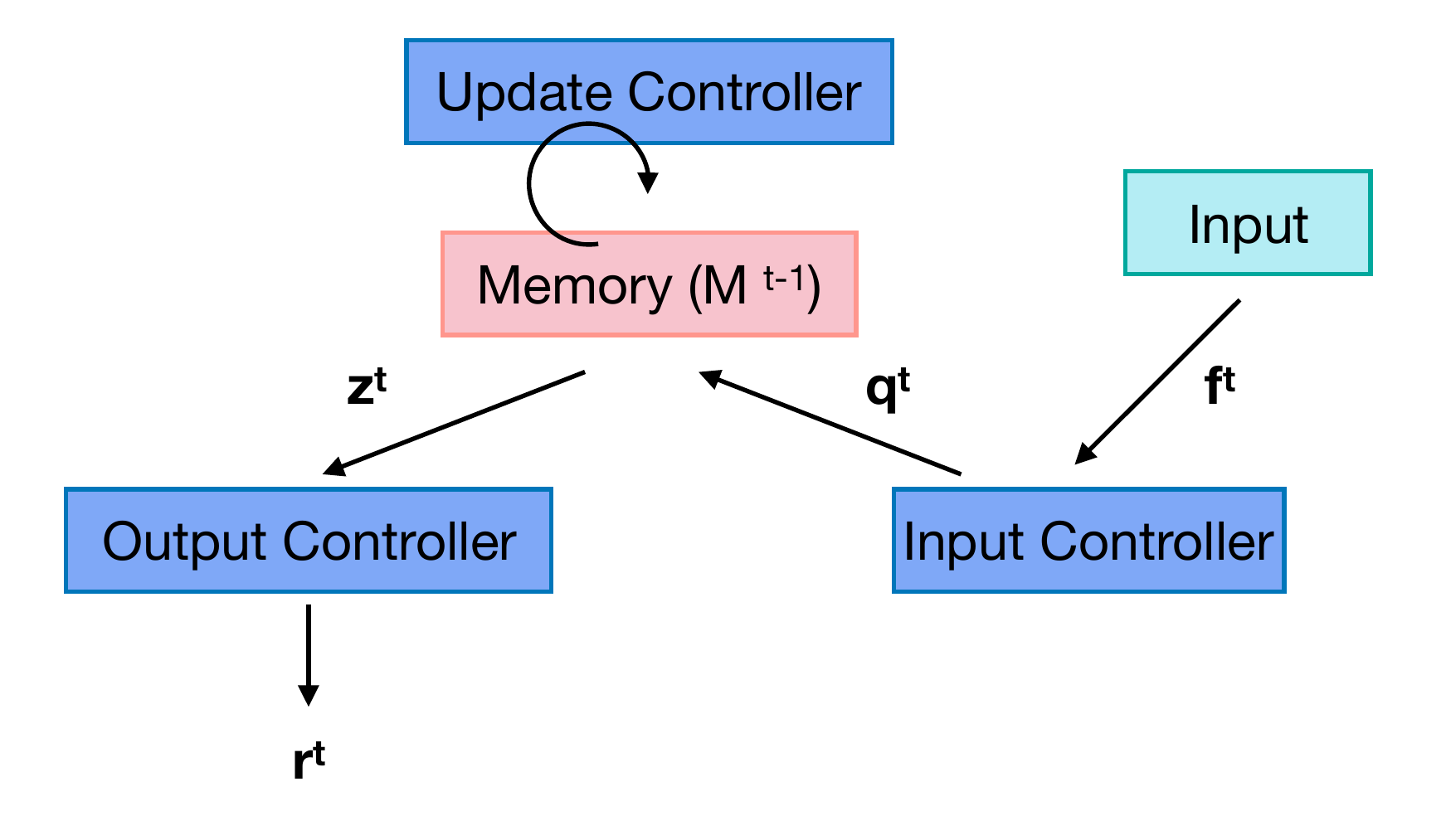}
\caption{Overview of an external memory which is composed of input, output and update controllers. The input controller determines what facts within the input are used to query the memory. The output controller determines what portion of the memory is passed as output. Finally the update controller updates the memory and propagates it to the next time step.}
\label{fig:lit_mem}
\end{figure}

This motivates the need for external memory components. Fig. \ref{fig:lit_mem} depicts the basic structure of a memory module. First, the input, $f$, is passed through an input controller which encodes the input and generates a query, $q$, to question the memory. The memory receives this query and using the facts stored in the current memory state, $M^{t-1}$, it synthesises the memory output, $z^t$, which is passed through an output controller to generate the memory output, $r^t$. The process finishes by updating the memory and propagating it to the next time step using an update controller. The update controller uses the current timestep's input $f^t$ and the memory output $z^t$ to generate the next memory state, $M^t$.

The use of external an memory is gaining traction in numerous domains, including language modelling \cite{munkhdalai2017neural}, visual question answering \cite{malinowski2014multi,xiong2016dynamic}, trajectory prediction \cite{fernando2018tree,fernando2018pedestrian} and reinforcement learning \cite{parisotto2017neural,FernandoTemporal}. In the seminal work of \cite{graves2014neural}, Graves et. al proposed a Neural Turing Machine (NTM) which uses this concept of an external memory to achieve superior results when recalling long-term dependencies and improved the generalisability of the learned representation, compared to conventional deep learned models like LSTMs. 

However, even with the augmented capacity offered by a memory network, the flat memory structure of  \cite{munkhdalai2017neural,xiong2016dynamic,malinowski2014multi,graves2014neural,sukhbaatar2015end,gemici2017generative,santoro2016meta} hinders the ability of the network to determine and exploit the long-term dependencies stored in memory. This structure ignores important historical behaviour which occurs only over long time periods, and is regularly observed in applications such as aircraft and pedestrian trajectory modelling \cite{fernando2018tree,fernando2018pedestrian}.

Authors in \cite{fernando2018tree,fernando2018pedestrian} demonstrated the value of a hierarchical memory structure over a flat structure when generating inferences. However, we observe multiple deficiencies in those architectures which restrict their utility for the face manipulation detection task. Firstly, even though the tree memory architectures of \cite{fernando2018tree,fernando2018pedestrian} are inherently hierarchical, they compress the information. In this process, no attention is paid to the current query. Hierarchical information compression is useful when extracting informative content from a myriad of facts stored in the memory, however, we speculate that different facts are informative under different contexts; and the hierarchical operation acts as a bottleneck for information flow.

Secondly, the tree structure mixes semantics from different input embeddings in the hierarchy. Hence, it is ineffective for faces as attributes from different faces should remain separate. In contrast, we propose a hierarchical memory architecture which preserves the identity of individual facts stored in the memory and effectively attends to those facts through a hierarchical attention mechanism to retrieve query specific information. 

\section{Hierarchical Memory Network Module}

The proposed Hierarchical Memory Network (HMN) follows a completely different approach to the current state-of-the-art face manipulation detection techniques such as \cite{cozzolino2018forensictransfer, sabir2019recurrent,rossler2018faceforensics,nguyen2019multi,peng2016optimized}. We utilise NMNs to produce long-term dependency mappings regarding the basic attributes of the observed face. Furthermore, in contrast to \cite{munkhdalai2017neural,xiong2016dynamic,malinowski2014multi,graves2014neural,fernando2018tree,fernando2018pedestrian} the proposed memory addressing mechanism preserves the identity of individual facts stored in the memory, and effectively attends to those facts through a hierarchical attention mechanism to retrieve query specific information. 

The overall architecture of the proposed HMN consists of an input module, an input level attention layer, and a memory module. These components are described in the following subsections. 

\subsection{Input Module}

\begin{figure}[htbp]
\centering
 \includegraphics[width=\linewidth]{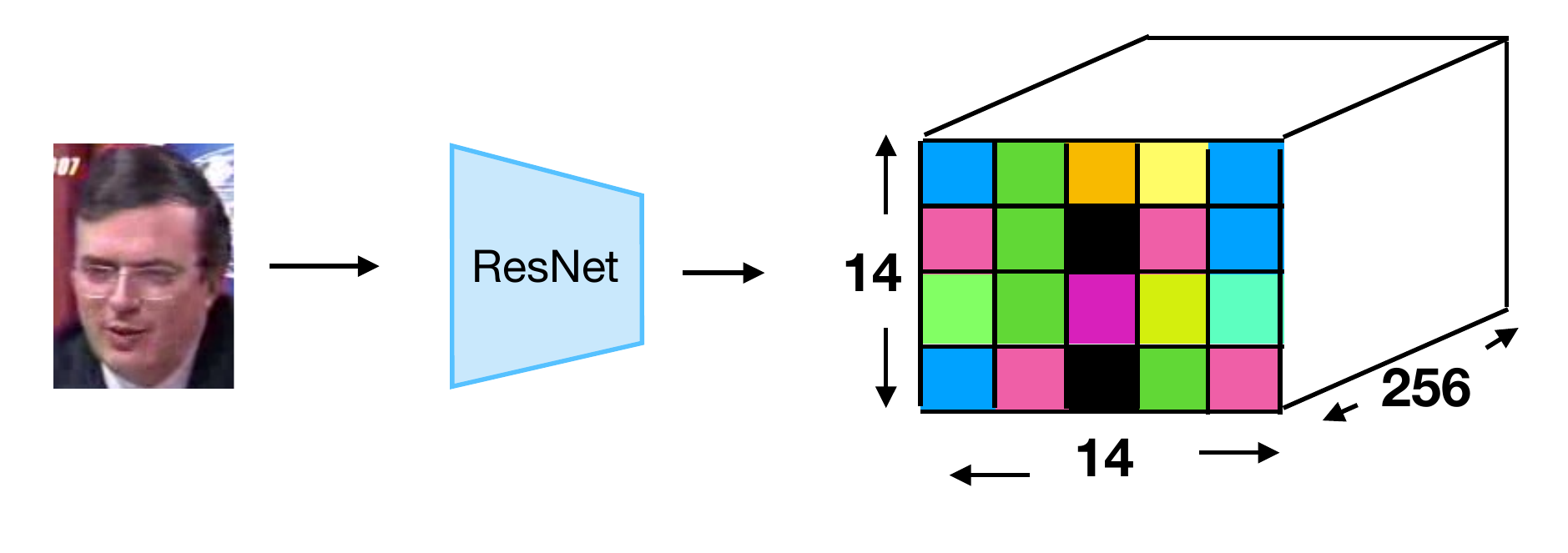}
\caption{Visual Feature Extraction: We utilise a ResNet \cite{he2016deep} CNN pre-trained on ImageNet \cite{deng2009imagenet} as our visual feature extractor, and extract features from the ``Activation-85'' layer with an output size of $14 \times 14 \times 256$.}
\label{fig:input}
\end{figure}

We extract features from input images using a pre-trained ResNet model trained on ImageNet. During pre-processing the input image is resized to $224 \times 224$ and we extract features from the ``Activation-85'' layer which has an output dimensionality, $d= 14 \times 14 \times 256$. Hence, as shown in Fig. \ref{fig:input}, the output has $14 \times 14= 196$ local patches, each containing $256$ features. The layer for feature extraction is evaluated experimentally, and this experiment is presented in Sec. \ref{sec:hyperpara}.

Formally, let the input face image $f^t$ at time step $t$ contain $k \in [1, K]$ patches. We summarise the information from neighbouring patches using a bidirectional GRU \cite{xiong2016dynamic}. In the forward pass, $\overrightarrow{f^{t}}$, of the GRU it reads $f^t$ from patches $1$ to $K$ while in the backward pass, $\overleftarrow{f^t}$, it reads from $K$ to $1$ using,
\begin{equation}
\begin{split}
\overrightarrow{f^{t}_k}= GRU_{fwd}(f_k^t, \overrightarrow{f_{k-1}^{t}}), \\
\overleftarrow{f^{t}_k}= GRU_{bwd}(f_k^t, \overleftarrow{f_{k+1}^{t}}),
\end{split}
\end{equation}
and concatenates the forward and backward vectors to generate a summary of the patch $k$,
\begin{equation}
\overleftrightarrow{f^t_k}= [\overrightarrow{f^{t}_k} ; \overleftarrow{f^{t}_k}].
\end{equation}

\subsection{Input Level Attention}
\label{sec:input_att}
Crucially, not all patches contribute equally when representing the salient attributes of a face. Hence, we utilise an attention mechanism that learns to pay varying levels of attention to patches when aggregating local patches into a single query vector, $q^t$. Specifically, we pass the encoded patches of the input image through a single layer MLP \cite{rumelhart1985learning} and obtain a representation, $v_k$ using,
\begin{equation}
v^t_k=\mathrm{tanh}(W_f\overleftrightarrow{f^t_k} + b_f),
\end{equation}
where $W_f$ and $b_f$ are the weight and bias of the MLP. Then we measure the importance of the current patch, $f_k^t$ using the similarity of $v^t_k$ with a patch level context vector, $v_f$. Following \cite{yang2016hierarchical} we randomly initialise this context vector and jointly learn it during training. The similarity score values are then normalised using a softmax function, 
\begin{equation}
\beta^t_k = \dfrac{\mathrm{exp}([v_k^t]^\top v_f)}{\sum_{k}\mathrm{exp}([v_k^t]^\top v_f)},
\label{eq:input_level_attention}
\end{equation}
generating a query vector to summarise the input image,
\begin{equation}
q^t=\sum_{k}\beta^t_k \overleftrightarrow{f^t_k}.
\end{equation}

\subsection{Memory Module}

\begin{figure*}[htbp]
\centering
 \includegraphics[width=.8\textwidth]{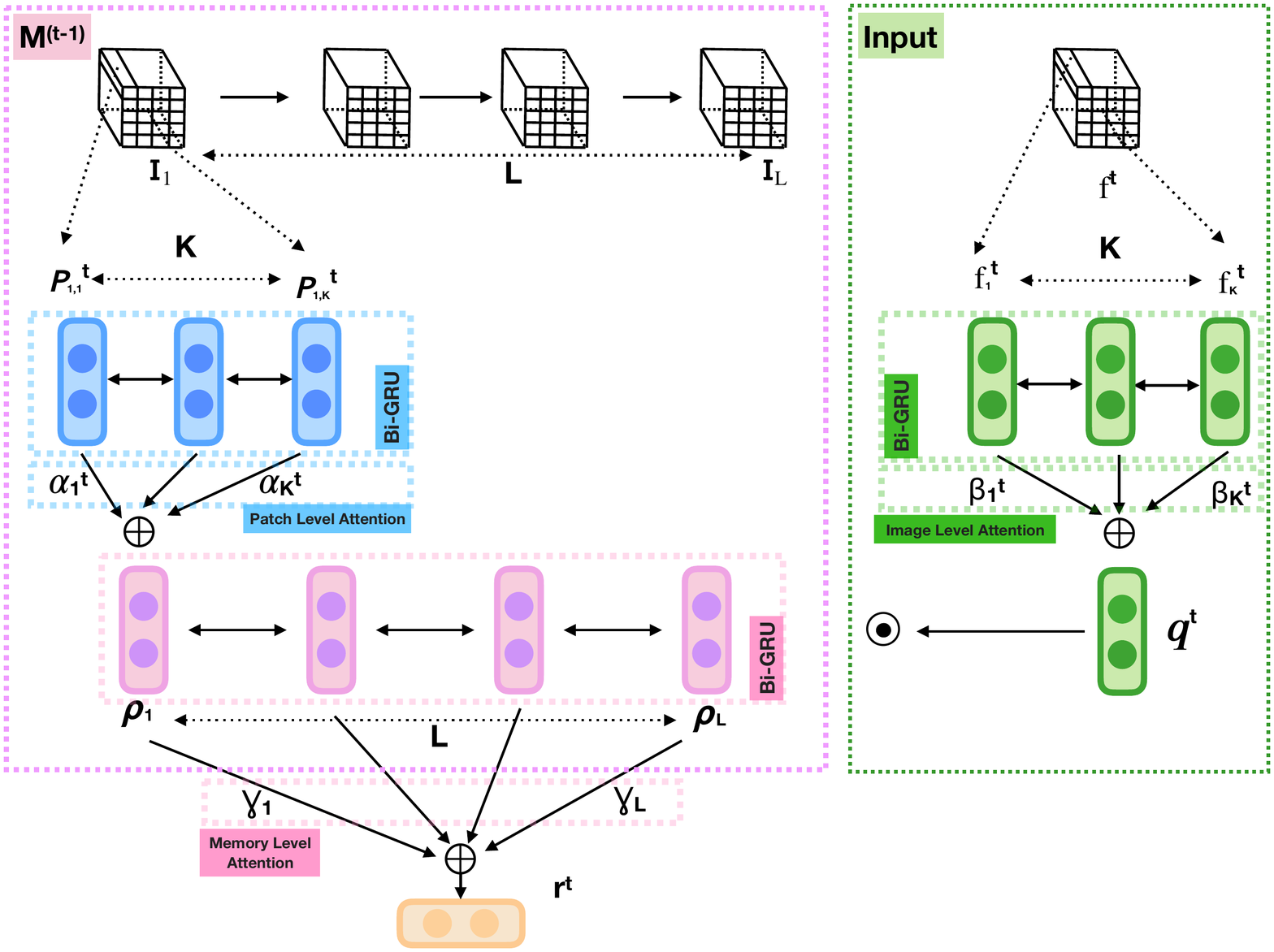}
\caption{Hierarchical Memory Module: Memory at time $t-1$ holds $L$ image embeddings (shown in the pink bounding box), each containing $K$ image patches (shown in blue). In the first level (patch level) we utilise attention, $\alpha$, which is computed over the incoming embedding (shown in green), to quantify the significance of each patch to the current query. In the second level of attention we attend to the memory content at the image level and extract informative components for decision making. Extracted information is propagated via the memory output, $r^t$, at time instant $t$. The overall memory output level attention, $\gamma_{i}$ where $i=[1..L]$, is a function of both the input embedding and the embeddings stored in memory.}
\label{fig:memory}
\end{figure*}

The proposed hierarchical memory module is illustrated in Fig. \ref{fig:memory}. Let the memory module, $M^{t-1}$, at time $t-1$ hold $L$ face embeddings, $I_i$, where $i \in [1, L]$ and each image contains $K$ patches, where $p_{i,k}$ is the $k^{th}$ patch of image $i$. 
\subsubsection{Patch Level Attention}
\label{sec:patch_att}
Following Sec. \ref{sec:input_att} we summarise each patch of an image that resides in the memory using,
\begin{equation}
\begin{split}
\overrightarrow{p_{i,k}} = GRU_{fwd}(p_{i,k}, \overrightarrow{p_{i,k-1}}), \\
\overleftarrow{p_{i,k}} = GRU_{bwd}(p_{i,k}, \overleftarrow{p_{i,k+1}}), \\
\overleftrightarrow{p_{i,k}} = [\overrightarrow{p_{i,k}} ; \overleftarrow{p_{i,k}}].
\end{split}
\label{eq:patch_level_attention}
\end{equation}

Motivated by \cite{xiong2016dynamic}, in order to measure the similarity between $p_{i,k}$ that resides in memory and the input image patch $f_k^t$, we first generate an augmented vector by multiplying the patches of $I_i$ with the equivalent patch in $f^t$, and concatenating this with the absolute difference between the patches,

\begin{algorithm} 
\begin{algorithmic}[1]

\ForEach {$i \in L $}
\ForEach {$k \in  K $}
\State $x_{i,k}=[\overleftrightarrow{p_{i,k}} \cdotp f_k^t ; |\overleftrightarrow{p_{i,k}} - f_k^t| ]$
\EndFor
\EndFor

\end{algorithmic}
\end{algorithm}

Then we employ attention to determine the informative patches,
\begin{equation}
\begin{split}
u_{i,k}=\mathrm{tanh}(W_xx_{i,k} + b_x), \\
\alpha_{i,k} = \dfrac{\mathrm{exp}([u_{i,k}]^\top u_x)}{\sum_{k}\mathrm{exp}([u_{i,k}]^\top u_x)}, \\
\rho_i = \sum_{k}\alpha_{i,k} x_{i,k},
\end{split}
\end{equation}
where $W_x$ and $b_x$ are the weights and bias of a seperate single layer MLP \cite{pal1992multilayer}, $u_x$ is a context vector which is randomly initialised and jointly learned during the training process, and $\alpha_{i,k} $ are the normalised score values quantifying the similarity between patch $f_k^t$ of the current input image and patch, $p_{i,k}$, of the image, $I_i$, that resides in memory. Drawing similarities to the way the human brain operates, this process measures how the attributes are similar with respect to past experiences. 

\subsubsection{Memory Output}
\label{sec:memout_att}
We apply another level of encoding with attention to summarise the similarity at the image level. Specifically, 
\begin{equation}
\begin{split}
\overrightarrow{\rho_{i}} = GRU_{fwd}(\rho_{i}, \overrightarrow{\rho_{i-1}}), \\
\overleftarrow{\rho_{i}} = GRU_{bwd}(\rho_{i}, \overleftarrow{\rho_{i+1}}), \\
\overleftrightarrow{\rho_{i}} = [\overrightarrow{\rho_{i}} ; \overleftarrow{\rho_{i}}].
\end{split}
 \end{equation}
 
Then we generate an augmented context vector using multiple interactions between the memory content and the query vector, $q^t$, such that,
\begin{equation}
 z_{i}=[\overleftrightarrow{\rho_{i}} \cdotp q^t ; \overleftrightarrow{\rho_{i}} \cdotp r^{t-1}; |\overleftrightarrow{\rho_{i}} - q^t| ]; \forall{ i \in L ,}
 \end{equation}
 where $r^{t-1}$ is the memory output at time step $t-1$. Now the output of the memory read operation $r^t$ at time step $t$ can be generated by,
\begin{equation}
\begin{split}
o_{i}=\mathrm{tanh}(W_zz_{i} + b_z), \\
\gamma_{i} = \dfrac{\mathrm{exp}([o_{i}]^\top o_z)}{\sum_{k}\mathrm{exp}([o_{i}]^\top o_z)}, \\
r^t = \sum_{k}\gamma_{i} z_{i}, 
\end{split}
\label{eq:memory_level_attention}
\end{equation}

\subsubsection{Memory Update}
In the update procedure we directly append the current input, $f^t$, to the previous memory state, $M^{t-1}$, and propagate the memory to the next time step using,
\begin{equation}
M^t =[M^{t-1}_{2:L}; f^{t}].
\label{eq:memory_update}
\end{equation}

where we remove the oldest entry when appending a new embedding, maintaining a constant size.

\subsubsection{Fake Face Classification and Future Face Semantic Embedding Regression}
When performing the input face classification, $\hat{y}^t$, (i.e real or manipulated face) we directly pass the memory output, $r^t$, through a single layer MLP,
\begin{equation}
\hat{y}^t= \mathrm{sotmax}(W_yr^t + b_y).
\end{equation}

For the generation of future face patches we need to synthesise a sequence of $K$ patches. When decoding this information from the memory output, $r^t$, we apply the same strategy that we applied when encoding the patch information into a vector representation. Specifically,
\begin{equation}
\begin{split}
\overrightarrow{h_{k}^t} = GRU_{fwd}(h_{k}^t, \overrightarrow{h_{k-1}^t}), \\
\overleftarrow{h_{k}^t} = GRU_{bwd}(h_{k}^t, \overleftarrow{h_{k+1}^t}), \\
\overleftrightarrow{h_{k}^t} = [\overrightarrow{h_{k}} ; \overleftarrow{h_{k}}],
\end{split}
 \end{equation}
 where $h_0^t= r^t$. Then the relavant future face embeddings are predicted by, 

\begin{equation}
\hat{\eta}^t_k= \mathrm{Relu}(W_{\eta}\overleftrightarrow{h_{k}^t}+ b_{\eta}).
\end{equation}

\section{Model Learning}
It is tedious to hand engineer a loss for future face embedding prediction. Hence, motivated by the ability of GANs to seamlessly learn a task-specific loss function in numerous domains  \cite{choi2018stargan,karras2017progressive, zhang2017deep,fernando2018task, gupta2018social,sadeghian2018sophie}, we employ a generative adversarial framework. 

GANs are comprised of two components, a Generator, $G$, and a Discriminator, $D$, where they partake in a two player game. Utilising a random noise vector, $z^t$, $G$ tries to synthesise realistic future face embeddings, $\hat{\eta}^t$ and tries to fool $D$. The discriminator tries to discriminate the synthesised future face embeddings, $\hat{\eta}^t$, from real examples, $\eta^t$. Hence the loss of this process is not hand engineered and the framework learns a custom loss for the task at hand. This objective can be written as, 

\begin{equation}
V= \mathop{min}_{G} \mathop{max}_{D} \hspace{1mm} \mathbb{E}( log D( \eta^t)) 
+ \hspace{1mm}  \mathbb{E}(log(1 - D(G(z^t)))).
\end{equation}

However, in this framework $G$ randomly synthesises future face embeddings without considering any information regarding the current state of the system. Hence, we draw our inspiration from a variant of GANs called the conditional GAN \cite{isola2017image}, where $G$ learns a conditional mapping from a random noise vector, $z^t$, and current memory output, $r^t$, to an output, $\hat{\eta}^t$: $G(z^t, r^t) \rightarrow \hat{\eta}^t$. This augmented objective, $\check{V}$, can be written as,

\begin{equation}
\check{V} =\mathop{min}_{G} \mathop{max}_{D} \hspace{1mm} \mathbb{E}( log D(r^t, \eta^t)) 
+ \hspace{1mm}  \mathbb{E}(log(1 - D(r^t, G(z^t, r^t)))).
\label{eq:c_gan}
\end{equation}

We couple the objective in Eq. \ref{eq:c_gan} with the fake face classification loss, allowing the model to jointly learn an objective that considers both tasks. In addition to this, and following common practice in the GAN literature \cite{isola2017image, zhang2017deep,choi2018stargan} we add an $L_2$ regularisation of the synthesised future embeddings to encourage $G$ to generate realistic future embeddings. Our final objective, $V^*$, can be defined as,
\begin{equation}
V^* = \check{V} + log(\hat{y}^t) +  \sum_{k} || \eta^t_k - \hat{\eta^t_k}||^2 .
\end{equation}

\section{Experiments}

In this section we provide the implementation details of the proposed architecture and apply it to detect digital forgeries in three widely used public benchmarks. 

\subsection{Datasets}
\subsubsection{FaceForensics Dataset}
The FaceForensics dataset \cite{rossler2018faceforensics} was collected from YouTube. Videos are a resolution of at least 480p and are tagged with ``face'',``newscaster'' or ``newsprogram''. For the generation of tampered faces the authors utilise the Face2Face method \cite{thies2016face2face} between two randomly chosen videos. The dataset contains 704 videos for training (364,256 images), 150 videos for validation (76,309 images), and 150 videos for testing (78,562 images). 

\subsubsection{FaceForensics$++$ dataset}
This dataset \cite{rossler2019faceforensics++} is an extended version of FaceForensics and contains face manipulations from FaceSwap \cite{faceswap} and DeepFakes \cite{deepfake_app}. FaceSwap \cite{faceswap} is a light weight editing tool which copies the face region from one image to another using face maker positions. The original videos are taken from youtube and manually screened to remove occlusions. The dataset consists of 1,000 original videos and 3,000 manipulated videos (1,000 for each category) from Face2Face, FaceSwap and DeepFake methods. Similar to \cite{rossler2019faceforensics++} we select 720 videos for training, 140 for validation and 140 for testing.

\subsubsection{FakeFace in the Wild (FFW) Dataset}
The FFW dataset \cite{khodabakhsh2018fake} is constructed using a set of public videos from YouTube, and contains a wide array of fake content generated through computer graphics, GANs, manual and automatic tampering techniques, and their combinations. Therefore, it provides an ideal setting to evaluate the generalisability of the proposed methodology under a diverse set of manipulations. Videos have a variable duration from 2-74 seconds and have at least 480p resolution. In addition to these 150 forged videos, the FFW dataset contains 150 real face videos from FaceForensics \cite{rossler2018faceforensics}.

\subsection{Implementation Details}
Following \cite{afchar2018mesonet}, we applied the Viola-Jones face detector \cite{viola2001rapid} to detect faces. 
We extract every $20^{th}$ frame and the respective frame 15 frames ahead as the input-output pair for the future frame prediction task. To balance the dataset, frames are selected for extraction such that an equal number of samples are extracted from each video. For all the GRUs we set the hidden state dimension to 300. Values for hyper-parameters memory length, $L = 200$; and the number of patches, $K = 196$; are evaluated experimentally and these experiments are presented in Sec. \ref{sec:hyperpara}. Furthermore, in Sec. \ref{sec:hyperpara} we evaluate the effect of the distance between the input-output pair on the fake face classification and future fame prediction accuracies. 

For comparisons we utilise three baseline memory modules in our evaluations, Neural Turing Machine (NTM) \cite{graves2014neural}, the Dynamic Memory Network (DMN) \cite{xiong2016dynamic} and Tree Memory Network (TMN) \cite{fernando2018tree}. For fair comparison we train these methods using the same ResNet features utilised by the proposed method and we set the LSTM hidden state dimension of the NTM, DMN and TMN modules to 300 and memory lengths to 200. The extraction depth of the TMN is evaluated experimentally and is set to 3. We train these memories to directly classify the input image using supervised learning and binary cross entropy loss. 

\subsection{Face Reenactment Detection Using FaceForensics Dataset}
\label{sec:faceforensics}
The ability of the proposed method to detect facial reenactments is measured using the FaceForensics dataset \cite{rossler2018faceforensics}. We strictly adhere to the author's guidelines when pre-processing the data and used the same training, testing and validation splits as \cite{rossler2018faceforensics}. Following \cite{nguyen2018capsule}, we report the classification results in terms of video and frame level accuracy. Video level classifications are obtained by aggregating frame level predictions over the entire video and obtaining the most voted class label.

\begin{table}[htbp]
\centering
\caption{Evaluation of facial reenactment detection at the video and frame levels on the FaceForensics dataset \cite{rossler2018faceforensics} (higher is better). Current state-of-the-art methods for fake face detection are shown with a pink background, baseline memory models are shown with a blue background and the proposed HMN method is shown with a white background.}
\resizebox{1\linewidth}{!}{
\begin{tabular}{|c|c|c|c|c|}
\hline
                         &                                                 & \multicolumn{3}{c|}{Accuracy for different compression levels}                                                           \\ \cline{3-5} 
\multirow{-2}{*}{Level}  & \multirow{-2}{*}{Method}                        & 0 (None)                               & 23 (Light)                             & 40 (Strong)                            \\ \hline
                         & \cellcolor[HTML]{ECD0D0}Meso-4   \cite{afchar2018mesonet}                  & \cellcolor[HTML]{ECD0D0}-              & \cellcolor[HTML]{ECD0D0}95.30          & \cellcolor[HTML]{ECD0D0}-              \\ \cline{2-5} 
                         & \cellcolor[HTML]{ECD0D0}MesoInception-4  \cite{afchar2018mesonet}       & \cellcolor[HTML]{ECD0D0}-              & \cellcolor[HTML]{ECD0D0}95.30          & \cellcolor[HTML]{ECD0D0}-              \\ \cline{2-5} 
                         & \cellcolor[HTML]{ECD0D0}CapsuleForensics \cite{nguyen2018capsule}         & \cellcolor[HTML]{ECD0D0}99.30          & \cellcolor[HTML]{ECD0D0}98.00          & \cellcolor[HTML]{ECD0D0}82.00          \\ \cline{2-5} 
                         & \cellcolor[HTML]{ECD0D0}CapsuleForensics- Noise  \cite{nguyen2018capsule}   & \cellcolor[HTML]{ECD0D0}99.30          & \cellcolor[HTML]{ECD0D0}96.00          & \cellcolor[HTML]{ECD0D0}83.30          \\ \cline{2-5} 
                         & \cellcolor[HTML]{ECF4FF}NTM    \cite{graves2014neural}                  & \cellcolor[HTML]{ECF4FF}78.30          & \cellcolor[HTML]{ECF4FF}73.20          & \cellcolor[HTML]{ECF4FF}73.90          \\ \cline{2-5} 
                         & \cellcolor[HTML]{ECF4FF}DMN  \cite{xiong2016dynamic}                     & \cellcolor[HTML]{ECF4FF}81.50          & \cellcolor[HTML]{ECF4FF}80.10          & \cellcolor[HTML]{ECF4FF}75.50          \\ \cline{2-5} 
                         & \cellcolor[HTML]{ECF4FF}Tree Memory   \cite{fernando2018tree}              & \cellcolor[HTML]{ECF4FF}85.40          & \cellcolor[HTML]{ECF4FF}83.20          & \cellcolor[HTML]{ECF4FF}80.10          \\ \cline{2-5} 
\multirow{-8}{*}{Video}  & \cellcolor[HTML]{FFFFFF}HMN                     & \cellcolor[HTML]{FFFFFF}\textbf{99.95} & \cellcolor[HTML]{FFFFFF}\textbf{99.61} & \cellcolor[HTML]{FFFFFF}\textbf{96.34} \\ \hline\hline
                         & \cellcolor[HTML]{ECD0D0}Meso-4    \cite{afchar2018mesonet}                    & \cellcolor[HTML]{ECD0D0}94.60          & \cellcolor[HTML]{ECD0D0}92.40          & \cellcolor[HTML]{ECD0D0}83.20          \\ \cline{2-5} 
                         & \cellcolor[HTML]{ECD0D0}MesoInception-4     \cite{afchar2018mesonet}     & \cellcolor[HTML]{ECD0D0}96.80          & \cellcolor[HTML]{ECD0D0}93.40          & \cellcolor[HTML]{ECD0D0}81.30          \\ \cline{2-5} 
                         & \cellcolor[HTML]{ECD0D0}Nguyen et al.   \cite{nguyen2018modular}             & \cellcolor[HTML]{ECD0D0}98.80          & \cellcolor[HTML]{ECD0D0}96.10          & \cellcolor[HTML]{ECD0D0}76.40          \\ \cline{2-5} 
                         & \cellcolor[HTML]{ECD0D0}CapsuleForensics   \cite{nguyen2018capsule}       & \cellcolor[HTML]{ECD0D0}99.13          & \cellcolor[HTML]{ECD0D0}97.13          & \cellcolor[HTML]{ECD0D0}81.20          \\ \cline{2-5} 
                         & \cellcolor[HTML]{ECD0D0}CapsuleForensics- Noise  \cite{nguyen2018capsule}  & \cellcolor[HTML]{ECD0D0}99.37          & \cellcolor[HTML]{ECD0D0}96.50          & \cellcolor[HTML]{ECD0D0}81.00          \\ \cline{2-5} 
                         & \cellcolor[HTML]{ECD0D0}Zhou et al    \cite{zhou2017two}             & \cellcolor[HTML]{ECD0D0}99.93          & \cellcolor[HTML]{ECD0D0}96.00          & \cellcolor[HTML]{ECD0D0}86.83          \\ \cline{2-5} 
                         & \cellcolor[HTML]{ECD0D0}Rossler et al.   \cite{rossler2018faceforensics}          & \cellcolor[HTML]{ECD0D0}99.93          & \cellcolor[HTML]{ECD0D0}98.13          & \cellcolor[HTML]{ECD0D0}87.81          \\ \cline{2-5} 
                         & \cellcolor[HTML]{ECF4FF}NTM    \cite{graves2014neural}                       & \cellcolor[HTML]{ECF4FF}81.54          & \cellcolor[HTML]{ECF4FF}75.19          & \cellcolor[HTML]{ECF4FF}72.28          \\ \cline{2-5} 
                         & \cellcolor[HTML]{ECF4FF}DMN  \cite{xiong2016dynamic}                       & \cellcolor[HTML]{ECF4FF}82.41          & \cellcolor[HTML]{ECF4FF}78.61          & \cellcolor[HTML]{ECF4FF}76.53          \\ \cline{2-5} 
                         & \cellcolor[HTML]{ECF4FF}Tree Memory      \cite{fernando2018tree}          & \cellcolor[HTML]{ECF4FF}85.45          & \cellcolor[HTML]{ECF4FF}82.53          & \cellcolor[HTML]{ECF4FF}80.50          \\ \cline{2-5} 
\multirow{-11}{*}{Frame} & \cellcolor[HTML]{FFFFFF}HMN                     & \cellcolor[HTML]{FFFFFF}\textbf{99.97} & \cellcolor[HTML]{FFFFFF}\textbf{99.65} & \cellcolor[HTML]{FFFFFF}\textbf{96.51} \\ \hline
\end{tabular}}
\label{tab:ff_1}
\end{table}

Facial reenactment detection evaluations on FaceForensics \cite{rossler2018faceforensics} at the video and frame level are presented in Tab. \ref{tab:ff_1}. When analysing the results it is clear that baseline system performance degrades significantly when the video compression level increases. This ratifies the observations presented in \cite{cozzolino2018forensictransfer,khodabakhsh2018fake} where the authors speculate that when the compression level increases the specific artefact that the system is focusing on degrades in clarity and significance. Hence, when the visual clarity degrades the CNN based tampered face detection systems such as Meso-4 \cite{afchar2018mesonet}, CapsuleForensics \cite{nguyen2018capsule}, Rossler et al. \cite{rossler2018faceforensics}, and Nguyen et al. \cite{nguyen2018modular} fail. For instance, in Tab. \ref{tab:ff_1}, for Rossler et al. \cite{rossler2018faceforensics} we observe a 12.12\% degradation of performance between no-compression and strong-compression classes. 
In contrast, the performance degradation between no-compression and strong-compression classes for the memory based models is not that significant. Even though the baseline DMN and TMN architectures fail to attain satisfactory performance due to there inheritant architectural deficiencies in memory structure, the performance difference between the compression classes is approximately ~5\%. We speculate from these results that our memory based models, which are comparing and contrasting the similarities between the observed facts stored in the memory, are not merely focusing on a specific artefact in the observed images (eg: compression artefacts) but mapping the overall structure of the face and the long-term dependencies between examples. 

Comparing the NTM, DMN and TMN memory architectures with the proposed memory architecture, we observe that they fail to achieve satisfactory performance due to inherit deficiencies. We speculate that the flat memory structure of NTM and DMN fails to propagate useful information to the output, while the TMN mixes the embeddings from different historical observations, making it difficult to discriminate the faces. In the proposed HMN model we rectify these deficiencies and further augment the performance through the joint learning of the future embeddings. Through evaluations in Sec. \ref{sec:importance_of_GAN} and \ref{sec:quality_of_predictions} we demonstrate that the two tasks complement each other. 

\subsection{Evaluations against different manipulation types using FaceForensics$++$}

We evaluate the robustness of the proposed method to different state-of-the-art face manipulation techniques using FaceForensics$++$. In Tab. \ref{tab:ffplus_1} we compare the accuracies of the systems when trained on all manipulation types together.

\begin{table}[htbp]
\centering
\caption{Evaluation against different manipulation types at the frame level on the FaceForensics$++$ dataset \cite{rossler2019faceforensics++} (higher is better). Current state-of-the-art methods for fake face detection are shown with a pink background, baseline memory models are shown with a blue background and the proposed HMN method is shown with a white background.}
\resizebox{1\linewidth}{!}{
\begin{tabular}{ccccl}
\cline{1-4}
\multicolumn{1}{|c|}{\textbf{Method}}                                 & \multicolumn{3}{c|}{\textbf{Accuracy for different compression levels}}                                                                                                                       &  \\ \cline{1-4}
\multicolumn{1}{|c|}{}                      & \multicolumn{1}{c|}{\textbf{0 (None)}}             & \multicolumn{1}{c|}{\textbf{23 (Light)}}           & \multicolumn{1}{c|}{\textbf{40 (Strong)}}          &  \\ \cline{1-4}
\multicolumn{1}{|c|}{\cellcolor[HTML]{ECD0D0}MesoInception-4 \cite{afchar2018mesonet}}         & \multicolumn{1}{c|}{\cellcolor[HTML]{ECD0D0}96.51} & \multicolumn{1}{c|}{\cellcolor[HTML]{ECD0D0}85.51} & \multicolumn{1}{c|}{\cellcolor[HTML]{ECD0D0}75.65} &  \\ \cline{1-4}
\multicolumn{1}{|c|}{\cellcolor[HTML]{ECD0D0}ResNet \cite{he2016deep}}         & \multicolumn{1}{c|}{\cellcolor[HTML]{ECD0D0}88.24} & \multicolumn{1}{c|}{\cellcolor[HTML]{ECD0D0}81.10} & \multicolumn{1}{c|}{\cellcolor[HTML]{ECD0D0}62.15} &  \\ \cline{1-4}
\multicolumn{1}{|l|}{\cellcolor[HTML]{ECD0D0}Cozzolino et al. \cite{cozzolino2017recasting}}           & \multicolumn{1}{c|}{\cellcolor[HTML]{ECD0D0}98.56} & \multicolumn{1}{c|}{\cellcolor[HTML]{ECD0D0}79.56} & \multicolumn{1}{c|}{\cellcolor[HTML]{ECD0D0}56.38} &  \\ \cline{1-4}
\multicolumn{1}{|c|}{\cellcolor[HTML]{ECD0D0}Xception \cite{chollet2017xception}}        & \multicolumn{1}{c|}{\cellcolor[HTML]{ECD0D0}99.41} & \multicolumn{1}{c|}{\cellcolor[HTML]{ECD0D0}97.53} & \multicolumn{1}{c|}{\cellcolor[HTML]{ECD0D0}85.49} &  \\ \cline{1-4}
\multicolumn{1}{|c|}{\cellcolor[HTML]{ECF4FF}NTM  \cite{graves2014neural}}                     & \multicolumn{1}{c|}{\cellcolor[HTML]{ECF4FF}78.32}      & \multicolumn{1}{c|}{\cellcolor[HTML]{ECF4FF}74.24}      & \multicolumn{1}{c|}{\cellcolor[HTML]{ECF4FF}72.40}      &  \\ \cline{1-4}
\multicolumn{1}{|c|}{\cellcolor[HTML]{ECF4FF}DMN  \cite{xiong2016dynamic}}                     & \multicolumn{1}{c|}{\cellcolor[HTML]{ECF4FF}80.25}      & \multicolumn{1}{c|}{\cellcolor[HTML]{ECF4FF}75.23}      & \multicolumn{1}{c|}{\cellcolor[HTML]{ECF4FF}73.04}      &  \\ \cline{1-4}
\multicolumn{1}{|c|}{\cellcolor[HTML]{ECF4FF}Tree Memory  \cite{fernando2018tree}}             & \multicolumn{1}{c|}{\cellcolor[HTML]{ECF4FF}82.13}      & \multicolumn{1}{c|}{\cellcolor[HTML]{ECF4FF}78.33}      & \multicolumn{1}{c|}{\cellcolor[HTML]{ECF4FF}73.14}      &  \\ \cline{1-4}
\multicolumn{1}{|c|}{\cellcolor[HTML]{FFFFFF}HMN}                     & \multicolumn{1}{c|}{\cellcolor[HTML]{FFFFFF}\textbf{99.43}}      & \multicolumn{1}{c|}{\cellcolor[HTML]{FFFFFF}\textbf{98.65}}      & \multicolumn{1}{c|}{\cellcolor[HTML]{FFFFFF}\textbf{97.02}}      &  \\ \cline{1-4}
\multicolumn{1}{l}{}                                                  & \multicolumn{1}{l}{}                               & \multicolumn{1}{l}{}                               & \multicolumn{1}{l}{}                               & 
\end{tabular}}
\label{tab:ffplus_1}
\end{table}

Similar to the evaluations on the FaceForensics dataset (Sec. \ref{sec:faceforensics}) we observe that the performance of the baselines degrades rapidly with the increase in compression; while the proposed method achieves consistent accuracy between the none, light and strong compression categories regardless of the additional DeepFake and FaceSwap manipulation categories.

Tab. \ref{tab:ffplus_2} reports the accuracies for models trained on FaceForensics presented in Sec. \ref{sec:faceforensics} (i.e trained only using Face2Face manipulation technique) tested with the unseen DeepFake and FaceSwap manipulation types. Following \cite{nguyen2019multi} we use light compression videos (quantisation = 23) for this experiment and reported accuracies are at the image level. 

\begin{table}[htbp]
\centering
\caption{Evaluation against unseen manipulation types at the frame level on the FaceForensics$++$ dataset \cite{rossler2019faceforensics++} using the model trained in Sec. \ref{sec:faceforensics} (higher is better). Current state-of-the-art methods for fake face detection are shown with a pink background, baseline memory models are shown with a blue background and the proposed HMN method is shown with a white background.}
\begin{tabular}{|c|c|c|}
\hline
                         & \multicolumn{2}{c|}{Accuracy} \\ \cline{2-3} 
\multirow{-2}{*}{Method} & DeepFake      & FaceSwap      \\ \hline
\rowcolor[HTML]{ECD0D0} 
ResNet    \cite{he2016deep}                &    43.10           &      38.19         \\ \hline
\rowcolor[HTML]{ECD0D0} 
Xception     \cite{chollet2017xception}            &   55.12            &    50.38           \\ \hline
\rowcolor[HTML]{ECD0D0} 
Cozzolino et al.  \cite{cozzolino2017recasting}       &   62.61            &   52.29   \\ \hline
\rowcolor[HTML]{ECD0D0} 
Nguyen et al.    \cite{nguyen2019multi}        &    52.32           &     54.07   \\ \hline
\rowcolor[HTML]{ECF4FF} 
NTM    \cite{graves2014neural}                  &   50.45            &     46.35          \\ \hline
\rowcolor[HTML]{ECF4FF} 
DMN       \cite{xiong2016dynamic}                &   50.92            &    47.13           \\ \hline
\rowcolor[HTML]{ECF4FF} 
TMN        \cite{fernando2018tree}                &  51.34             &   48.19            \\ \hline
\rowcolor[HTML]{FFFFFF} 
HMN                      &  \textbf{84.12}             &   \textbf{86.53}       \\ \hline
\end{tabular}
\label{tab:ffplus_2}
\end{table}

When analysing the results in Tab. \ref{tab:ffplus_2} it is clear that all baseline methods struggle to cope with the unseen attacks. The Xception \cite{chollet2017xception} model that achieves commendable accuracy with the known attacks presented in Tab. \ref{tab:ffplus_1} struggles to generalise to unknown attacks as the method is focusing on specific artefacts in the training data that are left by the process that creates the fake face, and not learning a generalisable representation that segregates the two classes. In contrast, the performance gap for the memory based systems under seen and unseen attack types is considerably lower. This clearly demonstrates the importance of long-term dependency modelling using memory architectures that can be use to discriminate between fake and real faces.

\subsection{Detecting Unknown Attacks using FakeFace in the Wild (FFW) Dataset}

Following the recommendation in \cite{khodabakhsh2018fake} we measure the classification accuracy in terms of Equal Error Rates (EER), Attack Presentation Classification Error Rate (APCER) \cite{afchar2018mesonet} and Bonafide Presentation Classification Error Rate (BPCER) \cite{afchar2018mesonet} under three evaluation settings: (i) TestSet-I where there are 1,500 real face and 1,500 fake face images tampered with known attacks; (ii) TestSet-II with 1,500 real and 1,500 fake face samples with unknown attacks; and (iii) TestSet-III which is comprised of 1,776 real and 1,576 fake faces generated using the FaceSwap and SwapMe applications proposed by \cite{zhou2017two}.

\begin{table}[htbp]
\centering
\caption{Performance on known fake faces from TestSet-I of FFW \cite{khodabakhsh2018fake}. We report APCER, BPCER and EER \cite{afchar2018mesonet} as performance metrics, (lower values are better). Current state-of-the-art methods for fake face detection are shown with a pink background, baseline memory models are shown with a blue background and the proposed HMN method is shown with a white background.}
\begin{tabular}{ccccl}
\cline{1-4}
\multicolumn{1}{|c|}{\textbf{Method}}                         & \multicolumn{1}{c|}{\textbf{APCER}}               & \multicolumn{1}{c|}{\textbf{BPCER}}               & \multicolumn{1}{c|}{\textbf{EER}}                 &  \\ \cline{1-4}
\multicolumn{1}{|c|}{\cellcolor[HTML]{ECD0D0}LBP \cite{ahonen2006face}}             & \multicolumn{1}{c|}{\cellcolor[HTML]{ECD0D0}3.80} & \multicolumn{1}{c|}{\cellcolor[HTML]{ECD0D0}2.87} & \multicolumn{1}{c|}{\cellcolor[HTML]{ECD0D0}3.33} &  \\ \cline{1-4}
\multicolumn{1}{|c|}{\cellcolor[HTML]{ECD0D0}AlexNet \cite{krizhevsky2012imagenet}}         & \multicolumn{1}{c|}{\cellcolor[HTML]{ECD0D0}7.80} & \multicolumn{1}{c|}{\cellcolor[HTML]{ECD0D0}1.73} & \multicolumn{1}{c|}{\cellcolor[HTML]{ECD0D0}3.73} &  \\ \cline{1-4}
\multicolumn{1}{|c|}{\cellcolor[HTML]{ECD0D0}VGG19 \cite{simonyan2014very}}           & \multicolumn{1}{c|}{\cellcolor[HTML]{ECD0D0}2.47} & \multicolumn{1}{c|}{\cellcolor[HTML]{ECD0D0}0.47} & \multicolumn{1}{c|}{\cellcolor[HTML]{ECD0D0}1.40} &  \\ \cline{1-4}
\multicolumn{1}{|c|}{\cellcolor[HTML]{ECD0D0}ResNet50 \cite{he2016deep}}        & \multicolumn{1}{c|}{\cellcolor[HTML]{ECD0D0}2.27} & \multicolumn{1}{c|}{\cellcolor[HTML]{ECD0D0}0.47} & \multicolumn{1}{c|}{\cellcolor[HTML]{ECD0D0}1.40} &  \\ \cline{1-4}
\multicolumn{1}{|c|}{\cellcolor[HTML]{ECD0D0}Xception \cite{chollet2017xception}}        & \multicolumn{1}{c|}{\cellcolor[HTML]{ECD0D0}2.47} & \multicolumn{1}{c|}{\cellcolor[HTML]{ECD0D0}0.13} & \multicolumn{1}{c|}{\cellcolor[HTML]{ECD0D0}1.07} &  \\ \cline{1-4}
\multicolumn{1}{|c|}{\cellcolor[HTML]{ECD0D0}Inception \cite{szegedy2015going}}       & \multicolumn{1}{c|}{\cellcolor[HTML]{ECD0D0}0.67} & \multicolumn{1}{c|}{\cellcolor[HTML]{ECD0D0}0.47} & \multicolumn{1}{c|}{\cellcolor[HTML]{ECD0D0}0.53} &  \\ \cline{1-4}
\multicolumn{1}{|c|}{\cellcolor[HTML]{ECD0D0}Meso-4 \cite{afchar2018mesonet}}          & \multicolumn{1}{c|}{\cellcolor[HTML]{ECD0D0}0.61}     & \multicolumn{1}{c|}{\cellcolor[HTML]{ECD0D0}0.59}     & \multicolumn{1}{c|}{\cellcolor[HTML]{ECD0D0}0.56}     &  \\ \cline{1-4}
\multicolumn{1}{|c|}{\cellcolor[HTML]{ECD0D0}MesoInception-4 \cite{afchar2018mesonet}} & \multicolumn{1}{c|}{\cellcolor[HTML]{ECD0D0}0.55}     & \multicolumn{1}{c|}{\cellcolor[HTML]{ECD0D0}0.56}     & \multicolumn{1}{c|}{\cellcolor[HTML]{ECD0D0}0.53}     &  \\ \cline{1-4}
\multicolumn{1}{|c|}{\cellcolor[HTML]{ECF4FF}NTM \cite{graves2014neural}}             & \multicolumn{1}{c|}{\cellcolor[HTML]{ECF4FF}1.55}     & \multicolumn{1}{c|}{\cellcolor[HTML]{ECF4FF}0.48}     & \multicolumn{1}{c|}{\cellcolor[HTML]{ECF4FF}1.98}     &  \\ \cline{1-4}
\multicolumn{1}{|c|}{\cellcolor[HTML]{ECF4FF}DMN \cite{xiong2016dynamic}}             & \multicolumn{1}{c|}{\cellcolor[HTML]{ECF4FF}1.44}     & \multicolumn{1}{c|}{\cellcolor[HTML]{ECF4FF}0.42}     & \multicolumn{1}{c|}{\cellcolor[HTML]{ECF4FF}1.98}     &  \\ \cline{1-4}
\multicolumn{1}{|c|}{\cellcolor[HTML]{ECF4FF}Tree Memory \cite{fernando2018tree}}     & \multicolumn{1}{c|}{\cellcolor[HTML]{ECF4FF}1.53}     & \multicolumn{1}{c|}{\cellcolor[HTML]{ECF4FF}0.23}     & \multicolumn{1}{c|}{\cellcolor[HTML]{ECF4FF}1.51}     &  \\ \cline{1-4}
\multicolumn{1}{|c|}{\cellcolor[HTML]{FFFFFF}HMN}             & \multicolumn{1}{c|}{\cellcolor[HTML]{FFFFFF}\textbf{0.12}}     & \multicolumn{1}{c|}{\cellcolor[HTML]{FFFFFF}\textbf{0.09}}     & \multicolumn{1}{c|}{\cellcolor[HTML]{FFFFFF}\textbf{0.10}}     &  \\ \cline{1-4}
\multicolumn{1}{l}{}                                          & \multicolumn{1}{l}{}                              & \multicolumn{1}{l}{}                              & \multicolumn{1}{l}{}                              & 
\end{tabular}
\label{tab:ffw_1}
\end{table}

\begin{table}[htbp]
\centering
\caption{Performance on unknown fake faces from TestSet-II of FFW \cite{khodabakhsh2018fake}. We report APCER, BPCER and EER \cite{afchar2018mesonet} as performance metrics, (lower values are better). Current state-of-the-art methods for fake face detection are shown with a pink background, baseline memory models are shown with a blue background and the proposed HMN method is shown with a white background.}
\begin{tabular}{ccccl}
\cline{1-4}
\multicolumn{1}{|c|}{\textbf{Method}}                         & \multicolumn{1}{c|}{\textbf{APCER}}                & \multicolumn{1}{c|}{\textbf{BPCER}}               & \multicolumn{1}{c|}{\textbf{EER}}                  &  \\ \cline{1-4}
\multicolumn{1}{|c|}{\cellcolor[HTML]{ECD0D0}LBP \cite{ahonen2006face}}             & \multicolumn{1}{c|}{\cellcolor[HTML]{ECD0D0}89.00} & \multicolumn{1}{c|}{\cellcolor[HTML]{ECD0D0}2.87} & \multicolumn{1}{c|}{\cellcolor[HTML]{ECD0D0}48.73} &  \\ \cline{1-4}
\multicolumn{1}{|c|}{\cellcolor[HTML]{ECD0D0}AlexNet \cite{krizhevsky2012imagenet}}         & \multicolumn{1}{c|}{\cellcolor[HTML]{ECD0D0}91.47} & \multicolumn{1}{c|}{\cellcolor[HTML]{ECD0D0}1.73} & \multicolumn{1}{c|}{\cellcolor[HTML]{ECD0D0}32.13} &  \\ \cline{1-4}
\multicolumn{1}{|c|}{\cellcolor[HTML]{ECD0D0}VGG19 \cite{simonyan2014very}}           & \multicolumn{1}{c|}{\cellcolor[HTML]{ECD0D0}90.73} & \multicolumn{1}{c|}{\cellcolor[HTML]{ECD0D0}0.47} & \multicolumn{1}{c|}{\cellcolor[HTML]{ECD0D0}29.40} &  \\ \cline{1-4}
\multicolumn{1}{|c|}{\cellcolor[HTML]{ECD0D0}ResNet50 \cite{he2016deep}}        & \multicolumn{1}{c|}{\cellcolor[HTML]{ECD0D0}89.53} & \multicolumn{1}{c|}{\cellcolor[HTML]{ECD0D0}0.47} & \multicolumn{1}{c|}{\cellcolor[HTML]{ECD0D0}30.33} &  \\ \cline{1-4}
\multicolumn{1}{|c|}{\cellcolor[HTML]{ECD0D0}Xception \cite{chollet2017xception}}        & \multicolumn{1}{c|}{\cellcolor[HTML]{ECD0D0}93.20} & \multicolumn{1}{c|}{\cellcolor[HTML]{ECD0D0}0.13} & \multicolumn{1}{c|}{\cellcolor[HTML]{ECD0D0}26.87} &  \\ \cline{1-4}
\multicolumn{1}{|c|}{\cellcolor[HTML]{ECD0D0}Inception \cite{szegedy2015going}}       & \multicolumn{1}{c|}{\cellcolor[HTML]{ECD0D0}91.93} & \multicolumn{1}{c|}{\cellcolor[HTML]{ECD0D0}0.47} & \multicolumn{1}{c|}{\cellcolor[HTML]{ECD0D0}27.47} &  \\ \cline{1-4}
\multicolumn{1}{|c|}{\cellcolor[HTML]{ECD0D0}Meso-4 \cite{afchar2018mesonet}}          & \multicolumn{1}{c|}{\cellcolor[HTML]{ECD0D0}93.90}      & \multicolumn{1}{c|}{\cellcolor[HTML]{ECD0D0}1.05}     & \multicolumn{1}{c|}{\cellcolor[HTML]{ECD0D0}31.13}      &  \\ \cline{1-4}
\multicolumn{1}{|c|}{\cellcolor[HTML]{ECD0D0}MesoInception-4 \cite{afchar2018mesonet}} & \multicolumn{1}{c|}{\cellcolor[HTML]{ECD0D0}93.71}      & \multicolumn{1}{c|}{\cellcolor[HTML]{ECD0D0}0.89}     & \multicolumn{1}{c|}{\cellcolor[HTML]{ECD0D0}29.10}      &  \\ \cline{1-4}
\multicolumn{1}{|c|}{\cellcolor[HTML]{ECF4FF}NTM \cite{graves2014neural}}             & \multicolumn{1}{c|}{\cellcolor[HTML]{ECF4FF}93.59}      & \multicolumn{1}{c|}{\cellcolor[HTML]{ECF4FF}2.94}     & \multicolumn{1}{c|}{\cellcolor[HTML]{ECF4FF}43.5}      &  \\ \cline{1-4}
\multicolumn{1}{|c|}{\cellcolor[HTML]{ECF4FF}DMN \cite{xiong2016dynamic}}             & \multicolumn{1}{c|}{\cellcolor[HTML]{ECF4FF}93.12}      & \multicolumn{1}{c|}{\cellcolor[HTML]{ECF4FF}2.92}     & \multicolumn{1}{c|}{\cellcolor[HTML]{ECF4FF}42.1}      &  \\ \cline{1-4}
\multicolumn{1}{|c|}{\cellcolor[HTML]{ECF4FF}Tree Memory \cite{fernando2018tree}}     & \multicolumn{1}{c|}{\cellcolor[HTML]{ECF4FF}88.62}      & \multicolumn{1}{c|}{\cellcolor[HTML]{ECF4FF}1.31}     & \multicolumn{1}{c|}{\cellcolor[HTML]{ECF4FF}34.10}      &  \\ \cline{1-4}
\multicolumn{1}{|c|}{\cellcolor[HTML]{FFFFFF}HMN}             & \multicolumn{1}{c|}{\cellcolor[HTML]{FFFFFF}\textbf{49.95}}      & \multicolumn{1}{c|}{\cellcolor[HTML]{FFFFFF}\textbf{0.12}}     & \multicolumn{1}{c|}{\cellcolor[HTML]{FFFFFF}\textbf{12.51}}      &  \\ \cline{1-4}
\multicolumn{1}{l}{}                                          & \multicolumn{1}{l}{}                               & \multicolumn{1}{l}{}                              & \multicolumn{1}{l}{}                               & 
\end{tabular}
\label{tab:ffw_2}
\end{table}

\begin{table}[htbp]
\centering
\caption{Performance on unknown fake faces from TestSet-III of FFW \cite{khodabakhsh2018fake}. We report APCER, BPCER and EER \cite{afchar2018mesonet} as performance metrics, (lower values are better). Current state-of-the-art methods for fake face detection are shown with a pink background, baseline memory models are shown with a blue background and the proposed HMN method is shown with a white background.}
\begin{tabular}{ccccl}
\cline{1-4}
\multicolumn{1}{|c|}{\textbf{Method}}                         & \multicolumn{1}{c|}{\textbf{APCER}}                & \multicolumn{1}{c|}{\textbf{BPCER}}                & \multicolumn{1}{c|}{\textbf{EER}}                  &  \\ \cline{1-4}
\multicolumn{1}{|c|}{\cellcolor[HTML]{ECD0D0}LBP \cite{ahonen2006face}}             & \multicolumn{1}{c|}{\cellcolor[HTML]{ECD0D0}90.16} & \multicolumn{1}{c|}{\cellcolor[HTML]{ECD0D0}3.43}  & \multicolumn{1}{c|}{\cellcolor[HTML]{ECD0D0}46.06} &  \\ \cline{1-4}
\multicolumn{1}{|c|}{\cellcolor[HTML]{ECD0D0}AlexNet \cite{krizhevsky2012imagenet}}         & \multicolumn{1}{c|}{\cellcolor[HTML]{ECD0D0}94.04} & \multicolumn{1}{c|}{\cellcolor[HTML]{ECD0D0}5.01}  & \multicolumn{1}{c|}{\cellcolor[HTML]{ECD0D0}43.03} &  \\ \cline{1-4}
\multicolumn{1}{|c|}{\cellcolor[HTML]{ECD0D0}VGG19 \cite{simonyan2014very}}           & \multicolumn{1}{c|}{\cellcolor[HTML]{ECD0D0}97.27} & \multicolumn{1}{c|}{\cellcolor[HTML]{ECD0D0}2.31}  & \multicolumn{1}{c|}{\cellcolor[HTML]{ECD0D0}44.93} &  \\ \cline{1-4}
\multicolumn{1}{|c|}{\cellcolor[HTML]{ECD0D0}ResNet50 \cite{he2016deep}}        & \multicolumn{1}{c|}{\cellcolor[HTML]{ECD0D0}89.40} & \multicolumn{1}{c|}{\cellcolor[HTML]{ECD0D0}8.22}  & \multicolumn{1}{c|}{\cellcolor[HTML]{ECD0D0}43.79} &  \\ \cline{1-4}
\multicolumn{1}{|c|}{\cellcolor[HTML]{ECD0D0}Xception \cite{chollet2017xception}}        & \multicolumn{1}{c|}{\cellcolor[HTML]{ECD0D0}93.15} & \multicolumn{1}{c|}{\cellcolor[HTML]{ECD0D0}3.43}  & \multicolumn{1}{c|}{\cellcolor[HTML]{ECD0D0}40.99} &  \\ \cline{1-4}
\multicolumn{1}{|c|}{\cellcolor[HTML]{ECD0D0}Inception \cite{szegedy2015going}}       & \multicolumn{1}{c|}{\cellcolor[HTML]{ECD0D0}71.64} & \multicolumn{1}{c|}{\cellcolor[HTML]{ECD0D0}22.58} & \multicolumn{1}{c|}{\cellcolor[HTML]{ECD0D0}46.39} &  \\ \cline{1-4}
\multicolumn{1}{|c|}{\cellcolor[HTML]{ECD0D0}Meso-4 \cite{afchar2018mesonet}}          & \multicolumn{1}{c|}{\cellcolor[HTML]{ECD0D0}75.34}      & \multicolumn{1}{c|}{\cellcolor[HTML]{ECD0D0}12.13}      & \multicolumn{1}{c|}{\cellcolor[HTML]{ECD0D0}45.12}      &  \\ \cline{1-4}
\multicolumn{1}{|c|}{\cellcolor[HTML]{ECD0D0}MesoInception-4 \cite{afchar2018mesonet}} & \multicolumn{1}{c|}{\cellcolor[HTML]{ECD0D0}73.51}      & \multicolumn{1}{c|}{\cellcolor[HTML]{ECD0D0}10.12}      & \multicolumn{1}{c|}{\cellcolor[HTML]{ECD0D0}43.10}      &  \\ \cline{1-4}
\multicolumn{1}{|c|}{\cellcolor[HTML]{ECF4FF}NTM \cite{graves2014neural}}             & \multicolumn{1}{c|}{\cellcolor[HTML]{ECF4FF}83.45}      & \multicolumn{1}{c|}{\cellcolor[HTML]{ECF4FF}14.93}      & \multicolumn{1}{c|}{\cellcolor[HTML]{ECF4FF}48.80}      &  \\ \cline{1-4}
\multicolumn{1}{|c|}{\cellcolor[HTML]{ECF4FF}DMN \cite{xiong2016dynamic}}             & \multicolumn{1}{c|}{\cellcolor[HTML]{ECF4FF}83.42}      & \multicolumn{1}{c|}{\cellcolor[HTML]{ECF4FF}14.95}      & \multicolumn{1}{c|}{\cellcolor[HTML]{ECF4FF}46.12}      &  \\ \cline{1-4}
\multicolumn{1}{|c|}{\cellcolor[HTML]{ECF4FF}Tree Memory \cite{fernando2018tree}}     & \multicolumn{1}{c|}{\cellcolor[HTML]{ECF4FF}79.09}      & \multicolumn{1}{c|}{\cellcolor[HTML]{ECF4FF}10.15}      & \multicolumn{1}{c|}{\cellcolor[HTML]{ECF4FF}43.72}      &  \\ \cline{1-4}
\multicolumn{1}{|c|}{\cellcolor[HTML]{FFFFFF}HMN}             & \multicolumn{1}{c|}{\cellcolor[HTML]{FFFFFF}\textbf{44.89}}      & \multicolumn{1}{c|}{\cellcolor[HTML]{FFFFFF}\textbf{1.51}}      & \multicolumn{1}{c|}{\cellcolor[HTML]{FFFFFF}\textbf{14.12}}      &  \\ \cline{1-4}
\multicolumn{1}{l}{}                                          & \multicolumn{1}{l}{}                               & \multicolumn{1}{l}{}                               & \multicolumn{1}{l}{}                               & 
\end{tabular}
\label{tab:ffw_3}
\end{table}

Following \cite{khodabakhsh2018fake} we perform comparative evaluations with respect to the widely used texture based LBP \cite{ahonen2006face} method, and state-of-the-art CNN architectures, AlexNet \cite{krizhevsky2012imagenet}, VGG19 \cite{simonyan2014very}, ResNet50 \cite{he2016deep}, Xception \cite{chollet2017xception}, and GoogleLetNet/InceptionV3 \cite{szegedy2015going}, and the popular MesoNet-4 and MesoInception-4 \cite{afchar2018mesonet} due to the public availability of their implementations. \footnote{We use the implementation provided by the authors and available at https://github.com/DariusAf/MesoNet}

When comparing Tab. \ref{tab:ffw_2} and \ref{tab:ffw_3} with the results presented in Tab. \ref{tab:ffw_1} it is clear that the performance of all the baseline systems in terms of APCER and EER drops significantly showing the lack of transferability of the baseline systems. For instance, we observe lower APCER for Meso-4 \cite{afchar2018mesonet} and MesoInception-4 \cite{afchar2018mesonet} architectures, compared to Inception \cite{szegedy2015going}  in Tab. \ref{tab:ffw_1}, but comparatively higher APCER in Tab. \ref{tab:ffw_2} and \ref{tab:ffw_3}. This clearly demonstrate that the state-of-the-art methods for fake face classification models are focusing on specific artefacts in the training data that are left by the process that creates the fake face, limiting their applicability. 

Considering the performance of the memory architectures, the worst performance observed is for the NTM model highlighting the inadequacy of a flat memory structure for propagating long-term semantic relationships. A slight performance boost is observed through the introduction of a tree-structured memory, however, due to the mixing of semantics between different embeddings that reside in the memory, this method fails to effectively learn to discriminate fake images from real. However, the performance degradation between seen and unseen attacks (in terms of APCER) is comparatively low compared to CNN based systems such as GoogleLetNet/InceptionV3 \cite{szegedy2015going}, and the popular MesoNet-4 and MesoInception-4 \cite{afchar2018mesonet}, supporting the notion that the long term dependency modelling of memory networks improves performance. Focusing on these properties, the proposed HMN method, by building theories about the observed face in terms of their appearance and their anticipated behaviour, has successfully gained an insight into the fundamentals of real human faces, allowing it to successfully discriminate real examples from fake. 

\section{Discussion}
\subsection{Ablation Study}
\subsubsection{Importance of Hierarchical Attention}
In order to evaluate the contribution of hierarchical attention within the proposed framework we conducted a series of ablation experiments where we remove attention at the input level ($\beta$) (see Sec. \ref{sec:input_att}), path level ($\alpha$) (see Sec. \ref{sec:patch_att}) and memory output level ($\gamma$) (see Sec \ref{sec:memout_att}). In these variants we directly aggregate the vectors into a single output vector. Evaluation results using FFW TestSet-III are presented in Tab. \ref{tab:importance_of_attention}.

\begin{table}[htbp]
\caption{Evaluation of the contribution of attention on FFW TestSet-III, (lower EER is better). Components after the ``/'' have been removed from the model.}
\centering
\begin{tabular}{|c|c|l}
\cline{1-2}
\textbf{Ablation Model}                                                      & \textbf{EER}          \\ \hline 
HMN / $\beta$                                                  &             20.72          \\ \hline 
HMN / $\alpha$                                                  &           23.17           \\ \hline 
HMN / $\gamma$                                                  &        31.91             \\ \hline 
HMN / ($\beta + \alpha$)                         &         29.33              \\ \hline 
HMN / ($\beta + \gamma$)                         &      33.44                 \\ \hline 
HMN / ($\alpha + \gamma$)                        &      32.91                 \\ \hline 
HMN / ($\alpha + \beta + \gamma$) &           38.12              \\ \hline 
\rowcolor[HTML]{FFFFFF} 
HMN                                                  & \textbf{14.12} \\ \hline 
\end{tabular}
\label{tab:importance_of_attention}
\end{table}

Firstly, the most significant contribution from attention is observed when encoding the memory output, as denoted by the highest degradation in performance. Secondly, we observe a significant impact when generating the image representation using patch level attention. This verifies our hypothesis that hierarchical attention is important when acquiring knowledge from stored memories. This is why even the models with individual attention levels, HMN / $\alpha$ and HMN / $\gamma$, haven't been able to obtain good performance. Thirdly, we observe a substantial contribution from $\beta$, which helps generate a query effectively from the current input. 

\subsubsection{Importance of the GAN learning Objective}
\label{sec:importance_of_GAN}
We evaluate the contribution of the GAN learning framework and the contribution from the future face prediction task when detecting fake faces. 

In the first ablation model, $\mathrm{HMN / GAN}$, we removed the GAN learning objective and trained it using supervised learning with a combination of classification loss and Mean Squared Error (MSE) between the predicted and ground truth future face semantics. Another ablation model, $\mathrm{HMN / (GAN + \eta)}$, is added where it removes the future face prediction task and directly optimises the fake face classification objective. It should be noted that both of these models do not utilise GAN learning.

\begin{table}[htbp]
\caption{Evaluation of the contribution of the GAN learning framework on FFW TestSet-III. We compare the proposed model with two ablation models in terms of EER, (lower EER is better). Components after the ``/'' have been removed from the model.}
\centering
\begin{tabular}{|c|c|l}
\cline{1-2}
\textbf{Ablation Model}                                                      & \textbf{EER}                \\ \hline
$\mathrm{HMN / (GAN + \eta)}$                                                  &          43.19                   \\ \hline 
$\mathrm{HMN / GAN}$                                                   &   41.52                          \\ \hline 
\rowcolor[HTML]{FFFFFF}  
HMN                                                  & \textbf{14.12}     \\ \hline 
\end{tabular}
\label{tab:importance_of_GAN}
\end{table}
From the results reported in Tab. \ref{tab:importance_of_GAN} we speculate that the future face classification task and GAN learning objective are both equally important. Even though we expect to obtain better performance in $\mathrm{HMN / (GAN + \eta)}$ without the future face embedding prediction overhead, this model suffers the highest degradation in performance. This provides evidence for our theory that understanding and anticipating future emotions and interactions about the perceived face helps detect unrealistic faces.  

\subsubsection{Importance of Multi-task Learning}
In this section we utilise the FaceForensics$++$ dataset \cite{rossler2019faceforensics++} and evaluate the contribution from the auxiliary task. Using the model trained on the FaceForensics dataset (Sec. \ref{sec:faceforensics}) we test the model using the test set of FaceForensics$++$ dataset's three fake face classes,  Face2Face which the model has already seen in the training data, and the unseen DeepFake and FaceSwap classes. The results are presented in Tabs. \ref{tab:auxiliary_task_1} and \ref{tab:auxiliary_task_2}, respectively. For comparisons we report the results observed by Nguyen et al. in \cite{nguyen2019multi} as well as different variants of the state-of-the-art NTM framework  \cite{graves2014neural}. 

We observe that our observations contradict the results of \cite{nguyen2019multi}, which found no significant contribution from the auxiliary task for unseen attack detection and a reduction in performance from the auxiliary task for the detection of seen attacks. In contrast we observe a substantial contribution from the proposed multi-task learning paradigm for both seen and unseen attacks. 

In addition, the comparisons with the state-of-the-art NTM framework \cite{graves2014neural} reveal that even when the method is augmented with the future semantic embedding anticipation auxiliary task and the GAN learning of the memory embeddings, the NTM fails to outperform the proposed HMN model due to the deficiencies of the memory structure. However, we observe an increase in the fake face detection accuracies in the $\mathrm{NTM + \eta}$ and $\mathrm{NTM + (GAN +\eta)}$ ablation models when the baseline NTM is augmented, highlighting the contributions of those individual components to our primary task.

\begin{table}[htbp]
\caption{Evaluation of the contribution of the multi-task learning on the seen attacks in terms of accuracy on FaceForensics$++$ dataset \cite{rossler2019faceforensics++}, (higher is better). Components after the ``/'' have been removed from the model while ``$+$'' denotes the addition of components.}
\centering
\begin{tabular}{|c|c|l}
\cline{1-2}
\textbf{Ablation Model}          & \textbf{Accuracy}      \\ \hline
\rowcolor[HTML]{ECD0D0}   
Nguyen et al. / Auxiliary-task \cite{nguyen2019multi} & 92.70	\\ \hline 
\rowcolor[HTML]{ECD0D0} 
Nguyen et al. \cite{nguyen2019multi} &	92.50	\\ \hline 
\rowcolor[HTML]{ECF4FF} 
$\mathrm{NTM} $  \cite{graves2014neural}  		& 72.40  \\ \hline
\rowcolor[HTML]{ECF4FF} 
$\mathrm{NTM + \eta}$    						&  74.15  \\ \hline
\rowcolor[HTML]{ECF4FF} 
$\mathrm{NTM + (GAN +\eta)}$    						&  76.81  \\ \hline 
\rowcolor[HTML]{FFFFFF} 
$\mathrm{HMN / (GAN + \eta)}$                                                 &       83.45          \\ \hline 
\rowcolor[HTML]{FFFFFF} 
$\mathrm{HMN / GAN}$                                                 &    85.32             \\ \hline 
\rowcolor[HTML]{FFFFFF} 
$\mathrm{HMN / (\alpha + \beta + \gamma)}$                                                 &    88.15             \\ \hline
\rowcolor[HTML]{FFFFFF} 
$\mathrm{HMN / \eta}$                                                 &      94.12           \\ \hline 
\rowcolor[HTML]{FFFFFF}     
HMN                                                  &  \textbf{99.43} \\ \hline
\end{tabular}
\label{tab:auxiliary_task_1}
\end{table}

\begin{table}[htbp]
\caption{Evaluation of the contribution of the multi-task learning on unseen attacks in terms of accuracy on FaceForensics$++$ dataset \cite{rossler2019faceforensics++}, (higher is better). Components after the ``/'' have been removed from the model while ``$+$'' denotes the addition of components.}
\centering
\begin{tabular}{|c|c|c|}
\hline
                         & \multicolumn{2}{c|}{Accuracy} \\ \cline{2-3} 
\multirow{-2}{*}{Method} & DeepFake      & FaceSwap      \\ \hline 
\rowcolor[HTML]{ECD0D0} 
Nguyen et al. / Auxiliary-task \cite{nguyen2019multi} & 51.21 & 53.39	\\ \hline 
\rowcolor[HTML]{ECD0D0} 
Nguyen et al. \cite{nguyen2019multi} &	52.21 &	54.07 \\ \hline 
\rowcolor[HTML]{ECF4FF} 
$\mathrm{NTM}$    	 \cite{graves2014neural}	&	 50.45            &     46.35  \\ \hline
\rowcolor[HTML]{ECF4FF} 
$\mathrm{NTM + \eta}$    						& 53.41 &  50.12 \\ \hline
\rowcolor[HTML]{ECF4FF} 
$\mathrm{NTM + (GAN +\eta)}$    						& 55.89 & 52.69 \\ \hline 
\rowcolor[HTML]{FFFFFF} 
$\mathrm{HMN / (GAN + \eta)}$                                                 &       61.78  &    63.44     \\ \hline  
\rowcolor[HTML]{FFFFFF} 
$\mathrm{HMN / GAN}$                                                 &    65.12       &  67.38    \\ \hline 
\rowcolor[HTML]{FFFFFF} 
$\mathrm{HMN / (\alpha + \beta + \gamma)}$                                                 &    70.23     &  72.78      \\ \hline
\rowcolor[HTML]{FFFFFF} 
$\mathrm{HMN / \eta}$                                                 &      78.03    & 80.54      \\ \hline 
\rowcolor[HTML]{FFFFFF}     
HMN                                                  &  \textbf{84.12} &  \textbf{86.53}\\ \hline
\end{tabular}
\label{tab:auxiliary_task_2}
\end{table}

\subsubsection{Discussion}
Through the results presented in Tabs. \ref{tab:importance_of_attention} - \ref{tab:auxiliary_task_2} we clearly illustrate that the performance gain of our approach is due to our three novel contributions: the introduction of a joint learning framework which is inspired by recent neuroscience findings and predicts future face embeddings as an auxiliary task; the novel hierarchical memory structure which utilises hierarchical attention for memory output generation and preserves the integrity of the stored embeddings; and automatic learning of a loss function for the multiple tasks at hand through the GAN learning framework. 

\begin{figure}
\centering
  \subfloat[][]{\includegraphics[width=.48\linewidth]{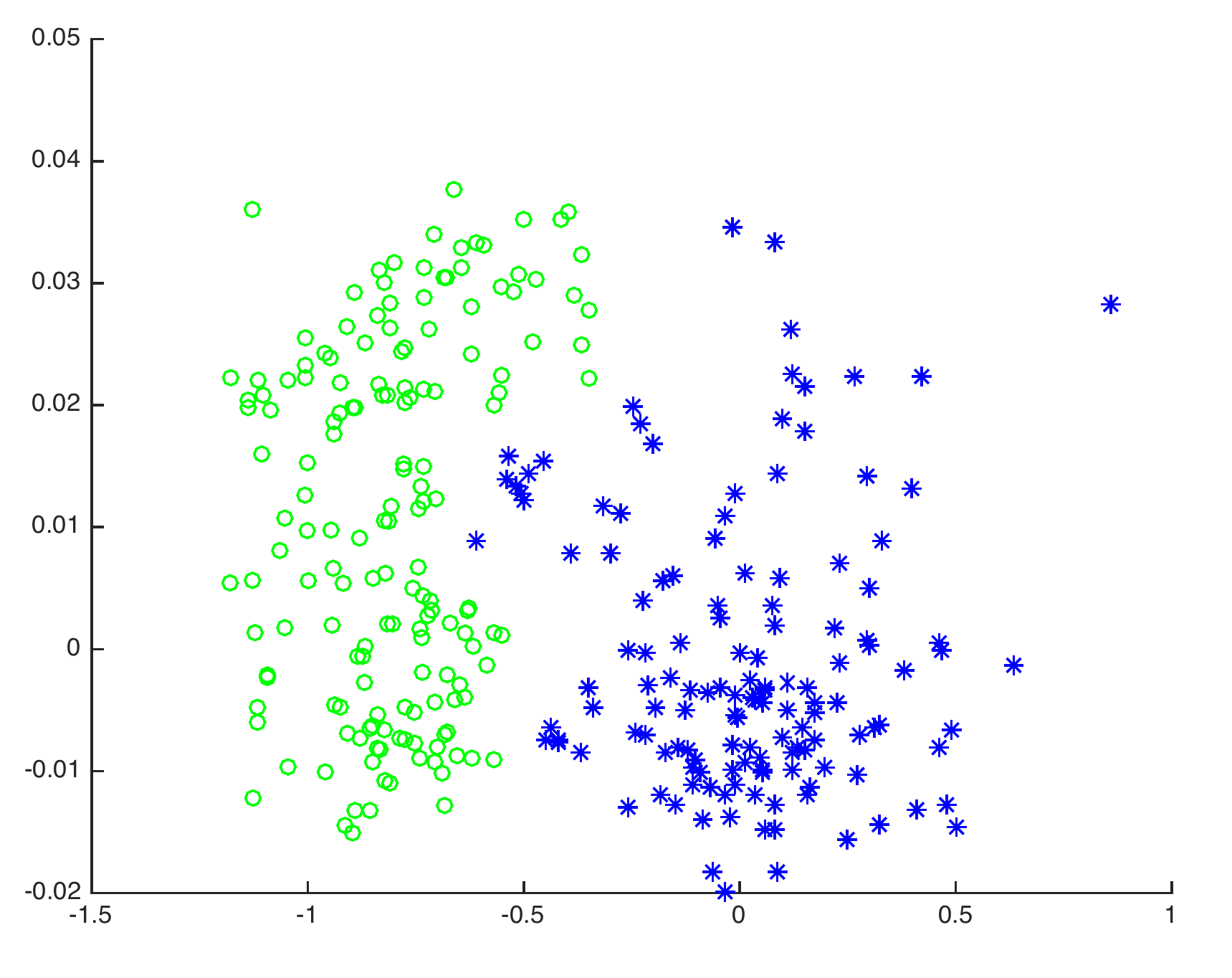}}
 \subfloat[][]{\includegraphics[width=.48\linewidth]{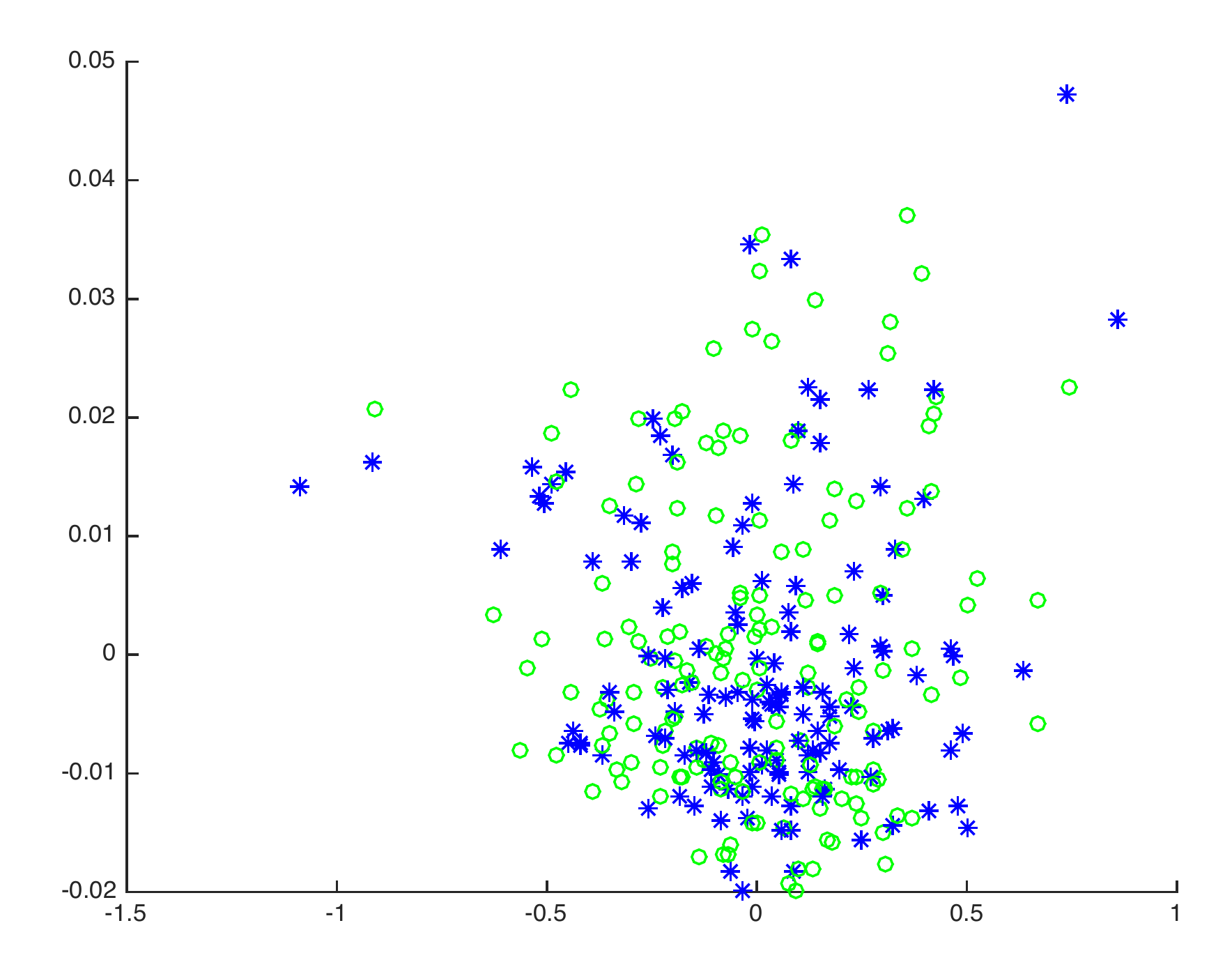}}
\caption{2D visualisation of the embedding space for the proposed HMN (a), and $\mathrm{HMN / \eta}$ ablation model presented in Tab. \ref{tab:auxiliary_task_2} (b) for 300 randomly selected images from the real and DeepFake classes in the FaceForensics$++$ dataset \cite{rossler2019faceforensics++}.}
\label{fig:embedding_space}
\end{figure}

To further demonstrate these merits, we visualise the embedding spaces for the HMN and $\mathrm{HMN / \eta}$ ablation model presented in Tab. \ref{tab:auxiliary_task_2}. We randomly selected 300 images from the real and DeepFake classes and extracted memory outputs $r^t$ from Eq. \ref{eq:memory_level_attention} for those inputs. We applied PCA \cite{jolliffe2011principal} to plot those embeddings in 2D, which are shown in Fig. \ref{fig:embedding_space}

When analysing these plots it is clear that through joint learning of both tasks the proposed HMN model has been able to learn a better separation between the real and fake faces compared to solely learning the single fake face classification task. 

\subsection{Quality of the Predicted Future Face Embeddings}
\label{sec:quality_of_predictions}
In order to understand the quality of the predicted semantic embeddings we measure the MSE between the predicted and ground truth embeddings on the FFW TestSet-III. For comparison we also report the MSE of the ablation model $\mathrm{HMN / GAN}$ introduced in Sec. \ref{sec:importance_of_GAN} and another ablation model, $\mathrm{HMN / y}$, where it uses GAN learning but only predicts the future face embeddings.

\begin{table}[htbp]
\caption{The quality of the predicted future face embeddings in terms of Mean Squared Error (MSE) on FFW TestSet-III, (lower EER is better). Components after the ``/'' have been removed from the model.}
\centering
\begin{tabular}{|c|c|l}
\cline{1-2}
\textbf{Ablation Model}                                                      & \textbf{MSE}      \\ \hline   
$\mathrm{HMN / GAN}$                                                 &    4.367             \\ \hline 
$\mathrm{HMN / y}$                                                 &    2.568             \\ \hline 
\rowcolor[HTML]{FFFFFF}     
HMN                                                  &  1.975\\ \hline
\end{tabular}
\label{tab:quality_of_futface}
\end{table}

When comparing the results in Tab. \ref{tab:quality_of_futface} we observe that more accurate predictions are generated by the proposed framework and that the GAN learning method has significantly contributed to alleviating errors in the embedding predictions. Furthermore, when comparing model $\mathrm{HMN / y}$ with the HMN, it is clear that the fake face classification objective has also contributed to improving the embedding prediction accuracy. 

\subsection{Effect of Memory Length, Number of Patches, Training Dataset Size and Distance Between Input-Output Pair}
\label{sec:hyperpara}
\begin{figure*}
\centering
 \subfloat[][Memory length, $L$ vs Accuracy]{\includegraphics[width=.3\textwidth]{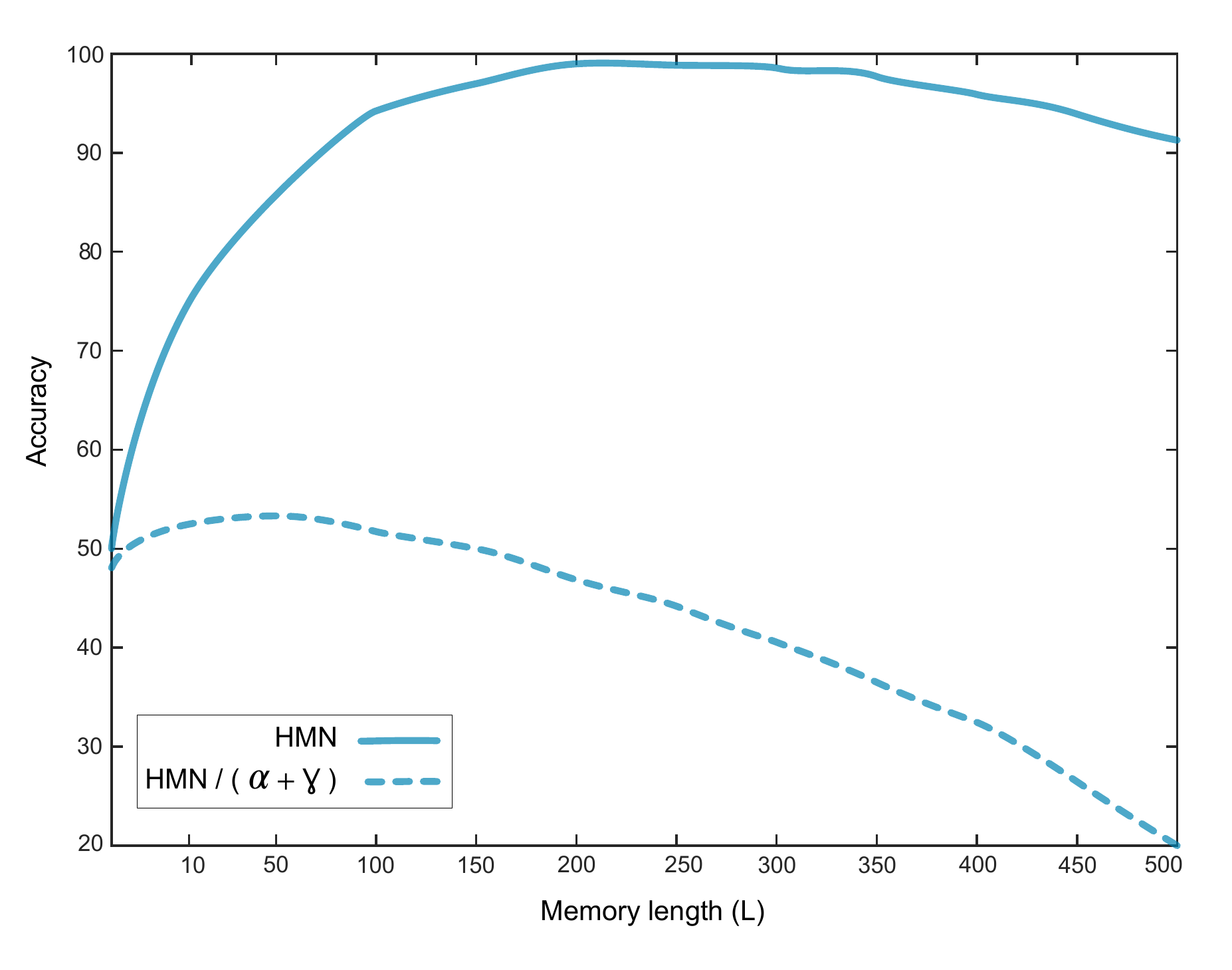}}
 \subfloat[][Number of patches, $K$ vs Accuracy]{\includegraphics[width=.3\textwidth]{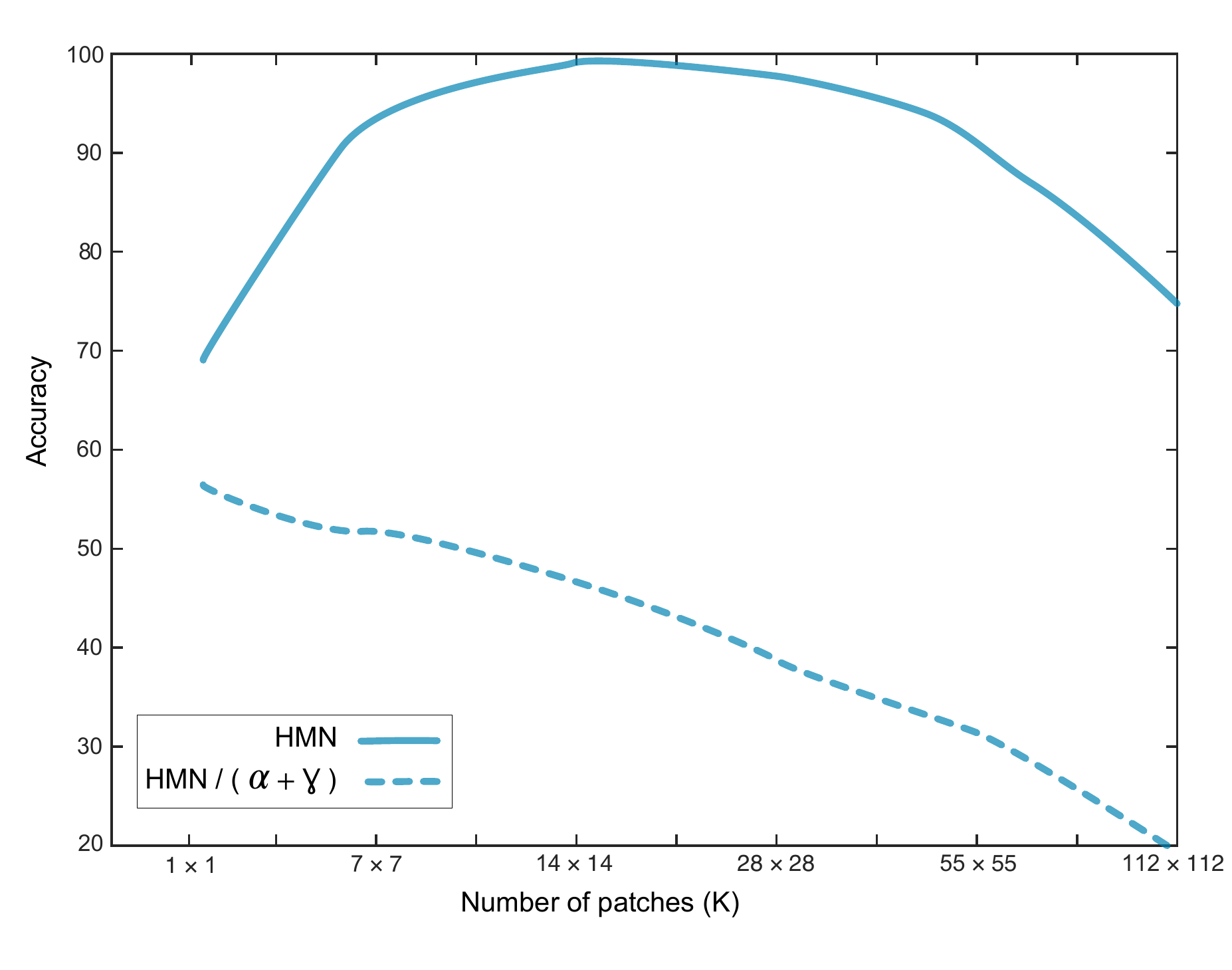}}
 \subfloat[][Training dataset size vs Accuracy]{\includegraphics[width=.3\textwidth]{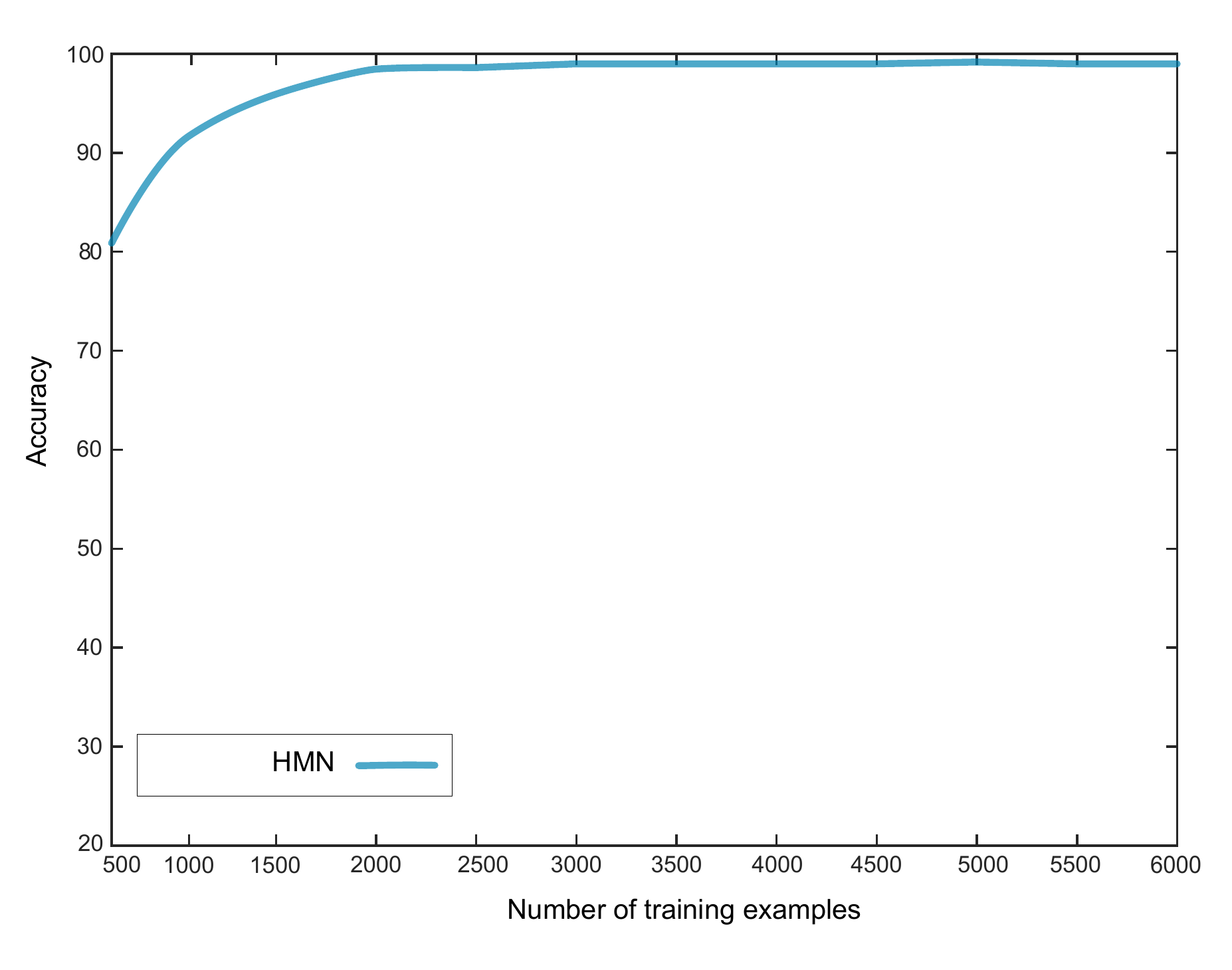}}
\caption{Hyper-parameter evaluation: We measure the effect of memory length, $L$, number of patches, $K$, and training dataset size against fake face detection accuracy.}
\label{fig:l_k}
\end{figure*}

In this experiment, we measure the change in accuracy with respect to memory length, $L$, the number of patches, $K$, that are extracted from the input image and the training dataset size in terms of images. In this experiment we use the validation split of the FaceForensics dataset \cite{rossler2018faceforensics}. In order to change the value $K$ we obtain features for different layers of ResNet, in particular we use the: activation\_50 ($112 \times 112$), activation\_59 ($55 \times 55$), activation\_71 ($28 \times 28$), activation\_85 ($14 \times 14$), activation\_98 ($7 \times 7$) and avg\_pool ($1 \times 1$) layers. Fig. \ref{fig:l_k} depicts these evaluations. It is observed that when adding more data to the memory the accuracy increases but converges around $L=200$. We observe a similar pattern with patch size, $K$, where it reaches a peak around $K=196$ (i.e $14 \times 14$). For comparison we also present these evaluations for the ablation model $HMN / (\alpha + \gamma)$. The model without attention fails to capture useful information when the data dimensionality increases, and performance continues to degrade as more information is added. By contrast, the proposed approach is able to leverage the additional data to improve performance. With Fig. \ref{fig:l_k} (c) we observe that a minimum of 3000 images are required to train the proposed framework. 

\begin{figure}
\centering
 \includegraphics[width=\linewidth]{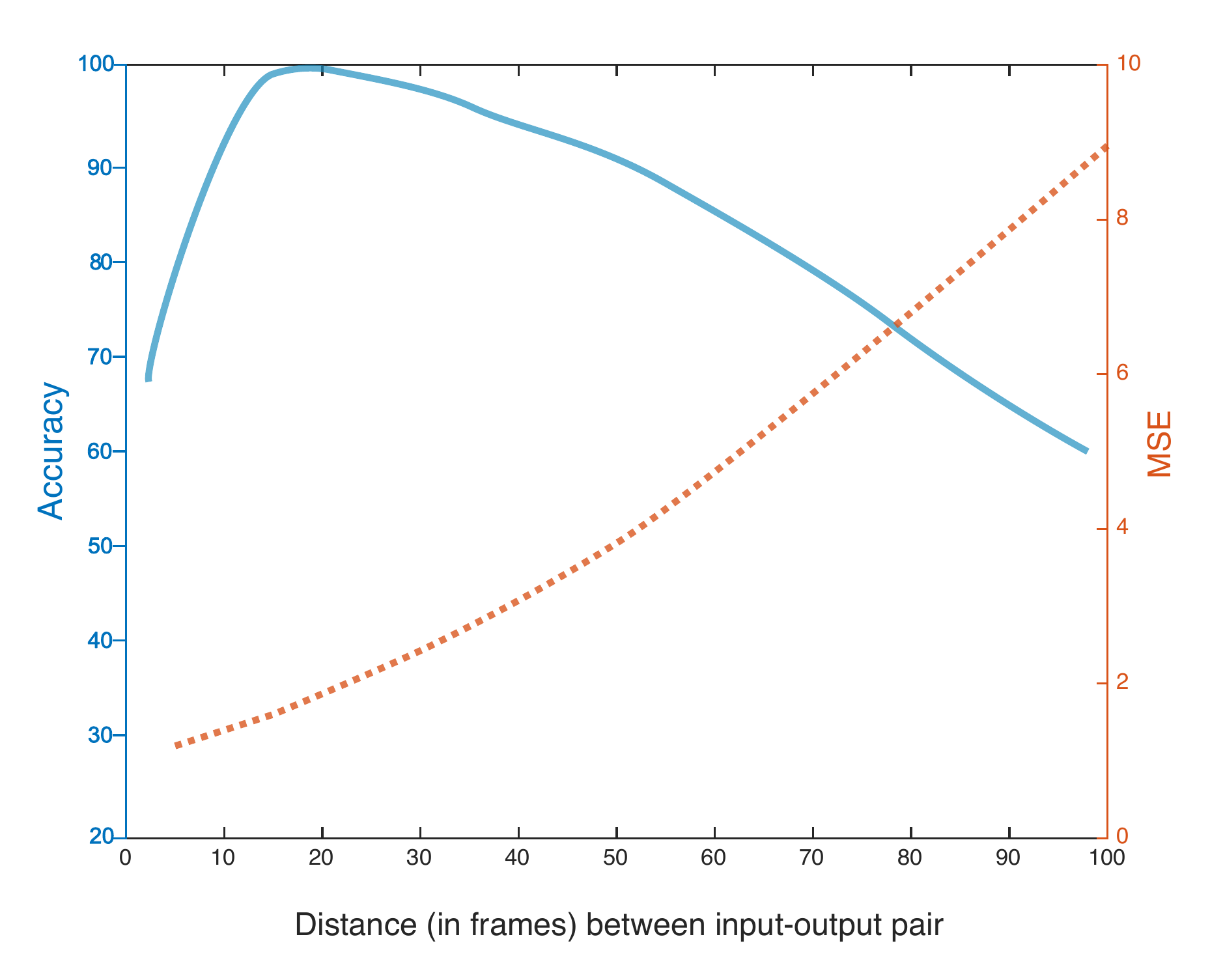}
\caption{Evaluation of distance (in frames) between the input-output pair for the future frame prediction task.}
\label{fig:distance_between_input_and_output}
\end{figure}

In an additional experiment, using the validation split of the FaceForensics dataset \cite{rossler2018faceforensics}, we measure the fake face classification accuracy and future embedding prediction error, in terms of MSE, for different sampling distances (in terms of frames) between the input-output pairs for the future frame prediction task. The results are presented in Fig. \ref{fig:distance_between_input_and_output}. We observe that the fake face classification accuracy as well as the quality of the future predicted embeddings degrades when the distance between the selected input output pairs increases. We speculate that the distance should be a generous size that allows the model learn to predict the future appearance and expressions of the observed face, but not too large as it becomes a nearly impossible task to perform and the facial information becomes uninformative for classifying the current observed face. 

\subsection{Hardware and Time Complexity Details}
The implementation of the HMN module presented in this paper is completed using Keras \cite{chollet2015keras} with a Theano \cite{bergstra2010theano} backend. The proposed HMN model does not require any special hardware such as GPUs to run and has 3M trainable parameters. We measured the time complexity of the proposed method using the test set of FaceForensics \cite{rossler2018faceforensics}. The proposed model is capable of generating 1000 predictions using 1000 $(14 \times 14 \times 256)$ input patches and generates 1000 image classifications and 1000 $(14 \times 14 \times 256)$ future face semantic predictions in 4021.2 seconds on a single core of an Intel Xeon E5-2680 2.50 GHz CPU. It should be noted that this prediction time includes the time taken to pass the test images through ResNet for embedding extraction, which takes 344.6 seconds.

\begin{figure}
\centering
  \subfloat[][]{\includegraphics[width=.48\linewidth]{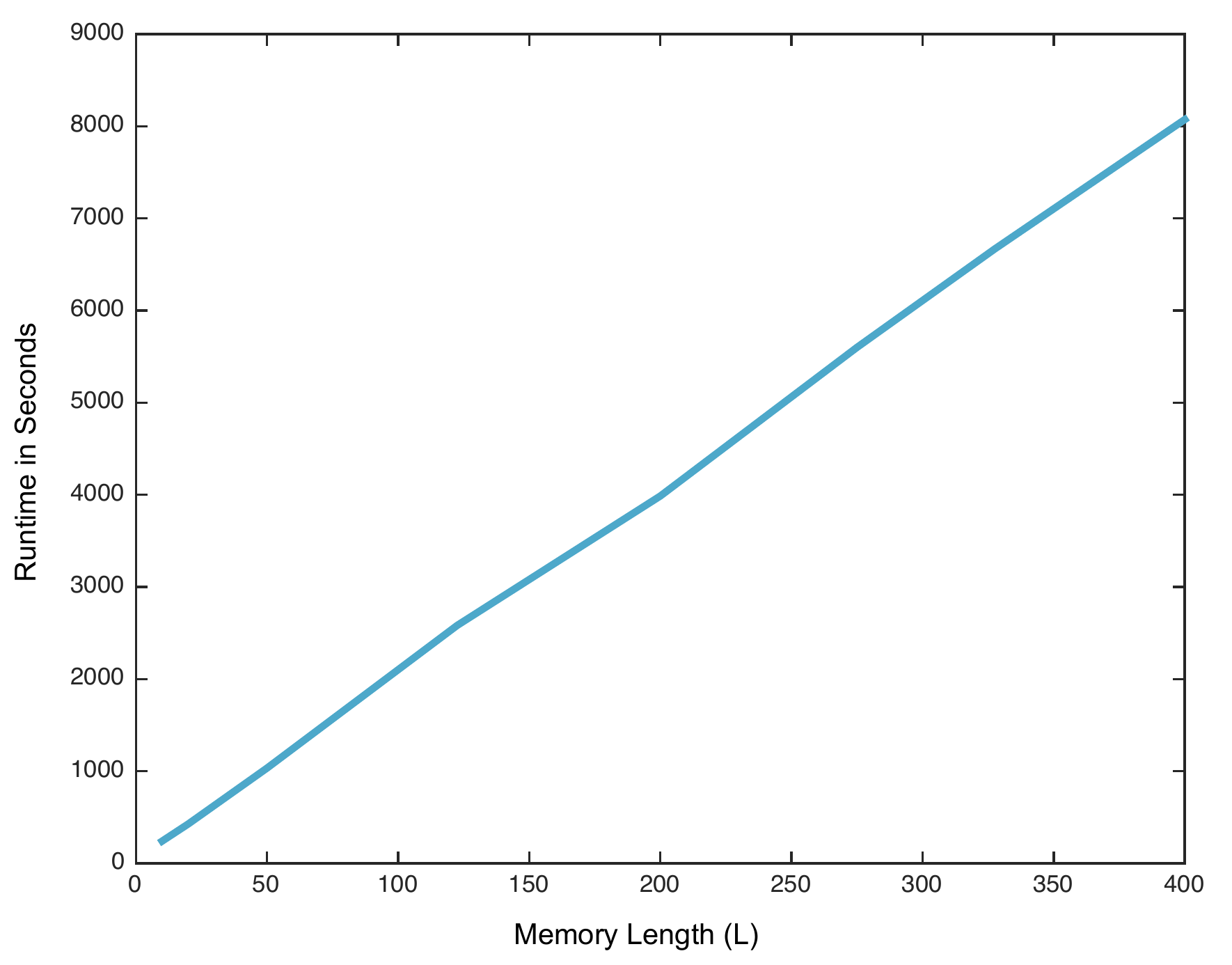}}
 \subfloat[][]{\includegraphics[width=.465\linewidth]{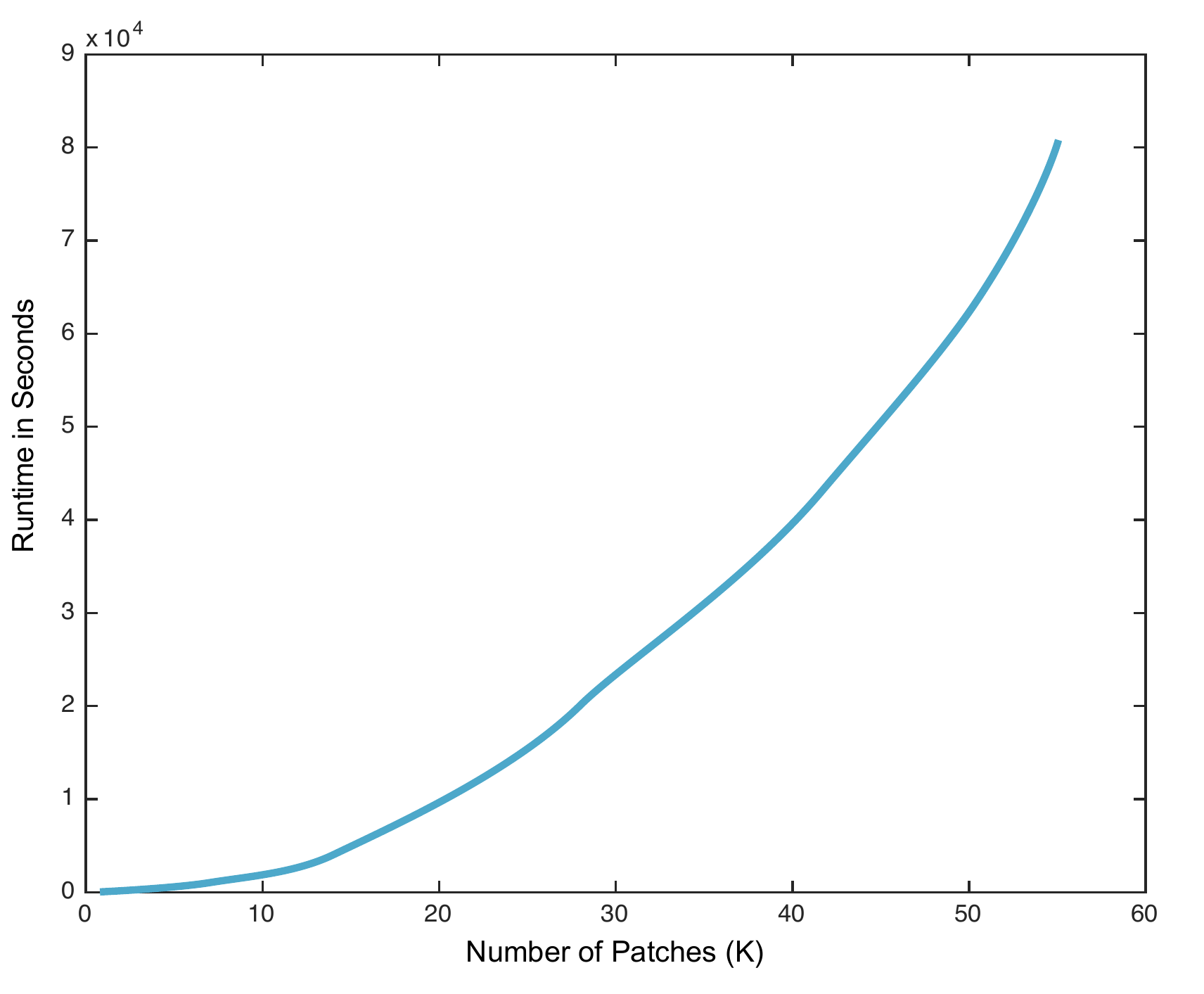}}
\caption{Evaluation of runtimes for different memory length, $L$ (a), and number of patches, $K$, values (b).}
\label{fig:runtimes}
\end{figure}

In the same experimental setting, we measured the time required to generate 1000 predictions for different lengths of the memory module, $L$, and different patch numbers, $K$. Results are given in Fig. \ref{fig:runtimes}. With the memory length the runtime grows approximately linearly, while with the number of patches it grows exponentially. This is because with the addition of each patch the hierarchical encoding and attention operations are utilised to accommodate the increasing dimensionality. 

\subsection{What is Actually Being Activated?}
\begin{figure*}
\centering
\includegraphics[width=\textwidth]{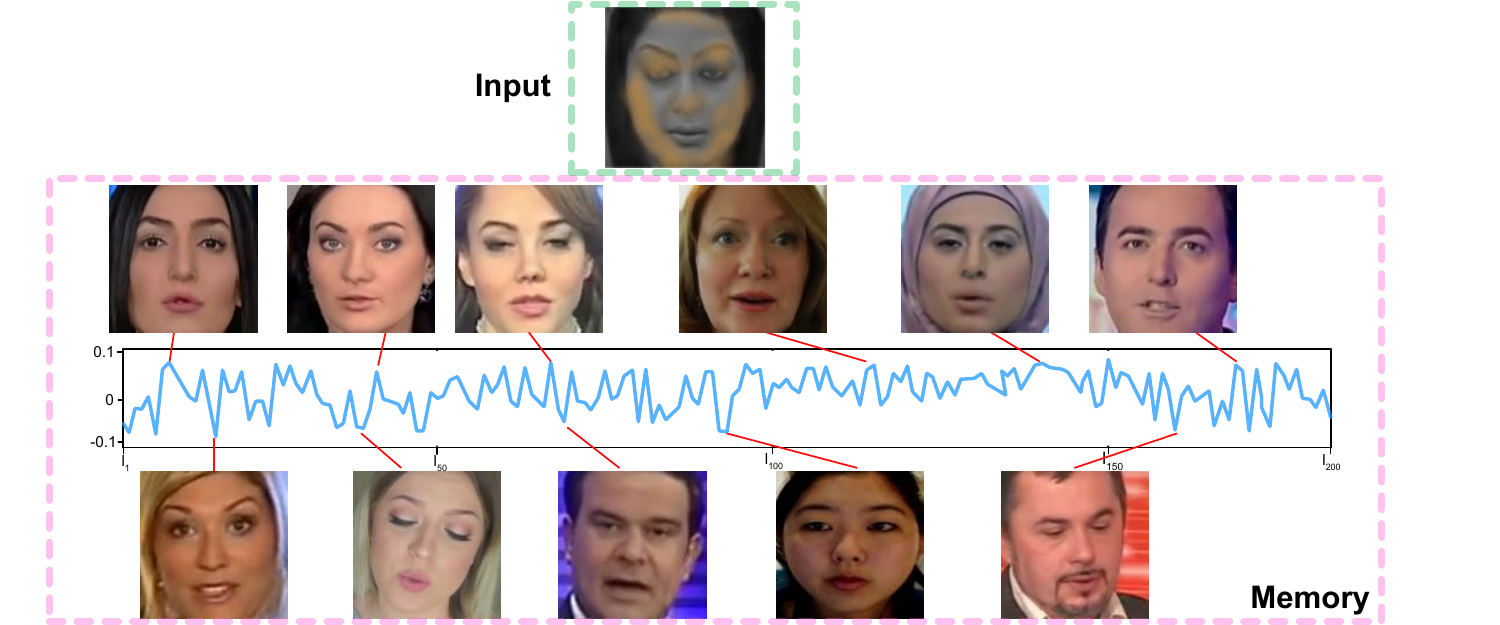}
\caption{Visualisation of memory activation, $\gamma$, when presented with the sample input shown at the top of the figure. For different peaks and valleys in the memory activation we also visualise what is stored in that particular memory location. The input level attention $\beta$ is interpolated and overlayed in yellow on top of the input. Brighter intensity values correspond to higher activations. Note that we have displayed the input image in grayscale for the clarity of the overlaid activations.}
\label{fig:activation_plot}
\end{figure*}

In Fig. \ref{fig:activation_plot} we visualise the $\gamma$ activation values, resulting from Eq. \ref{eq:memory_level_attention}, for the content of the memory for a sample input. We have overlayed the input level attention $\beta$, in yellow, on input (a brighter intensity corresponds to higher activations). As $L=200$ there exist 200 memory slots which we denote $l_1$ to $l_{200}$. For different peaks and valleys in the memory activation, we also show what input image embeddings are stored at that particular memory index. 

We observe that the HMN provides higher responses for similar face attribute patterns that the model has seen in the long-term history. It should be noted that even though the model hasn't seen the input image before, it is trying to anticipate the future behaviour while capturing the underlying semantics such as the movement of the eye brows, nose and head pose (see peaks between $l_{50}$ to $l_{150}$). The HMN model is measuring how much the current input is similar to the individual observations stored in memory by comparing and contrasting the fundamental facial attributes, allowing it to detect fake faces. This also highlights the importance of preserving the attributes of individual images separately in the proposed patch encoding mechanism. If this is not done (i.e. such as in the TMN) performance drops as the attributes of different faces are mixed. 

\begin{figure}
\centering
 \subfloat[][]{\includegraphics[width=.23\linewidth]{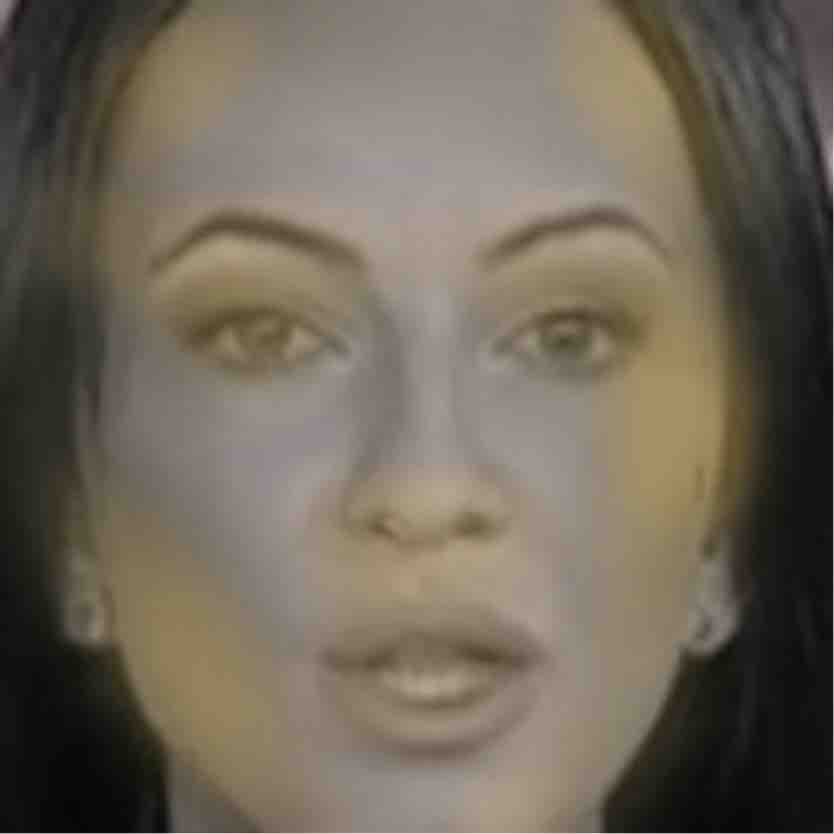}}
 \subfloat[][]{\includegraphics[width=.23\linewidth]{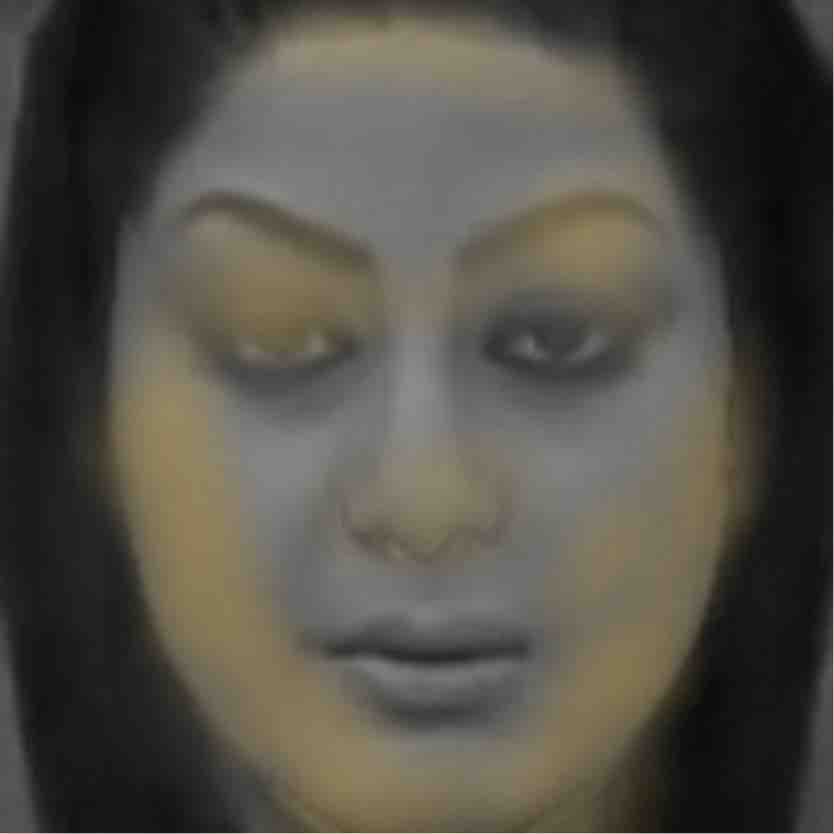}}
  \subfloat[][]{\includegraphics[width=.23\linewidth]{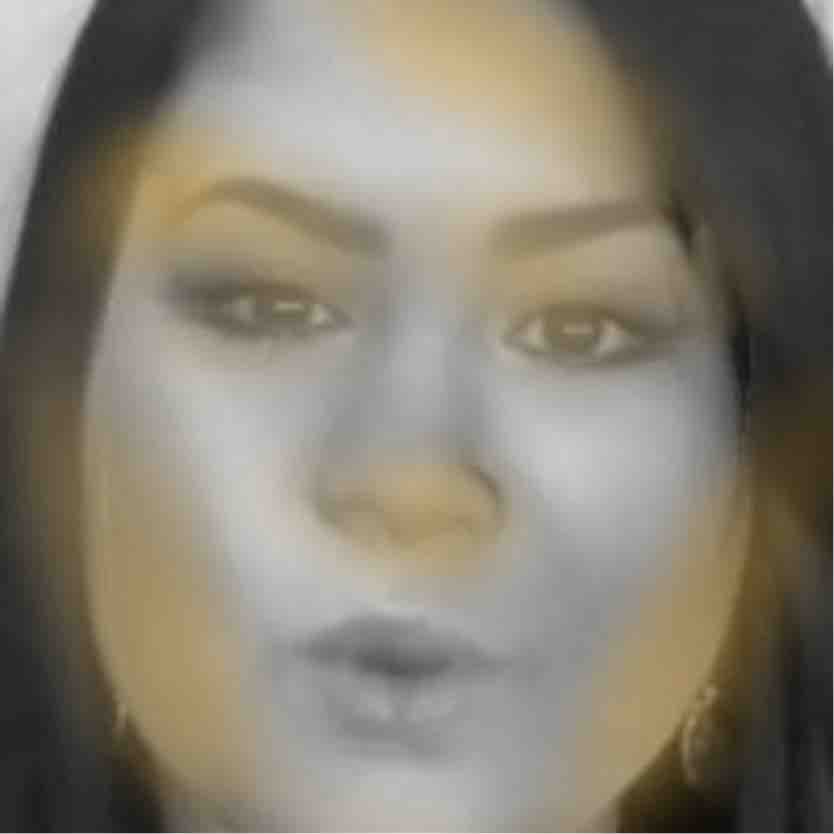}}
 \subfloat[][]{\includegraphics[width=.23\linewidth]{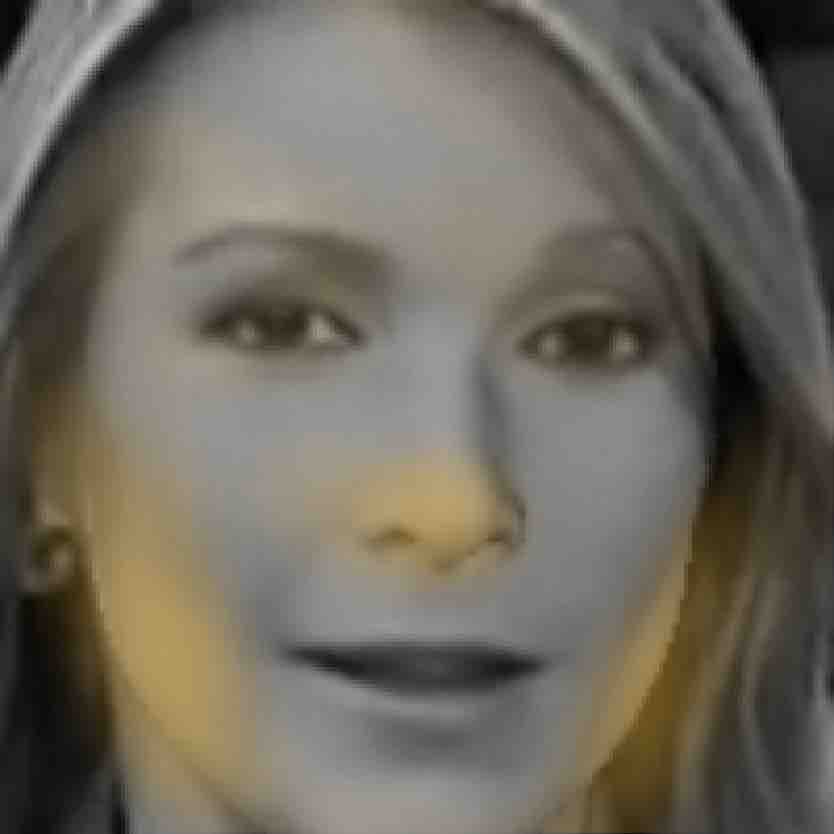}} \\
  \subfloat[][]{\includegraphics[width=.23\linewidth]{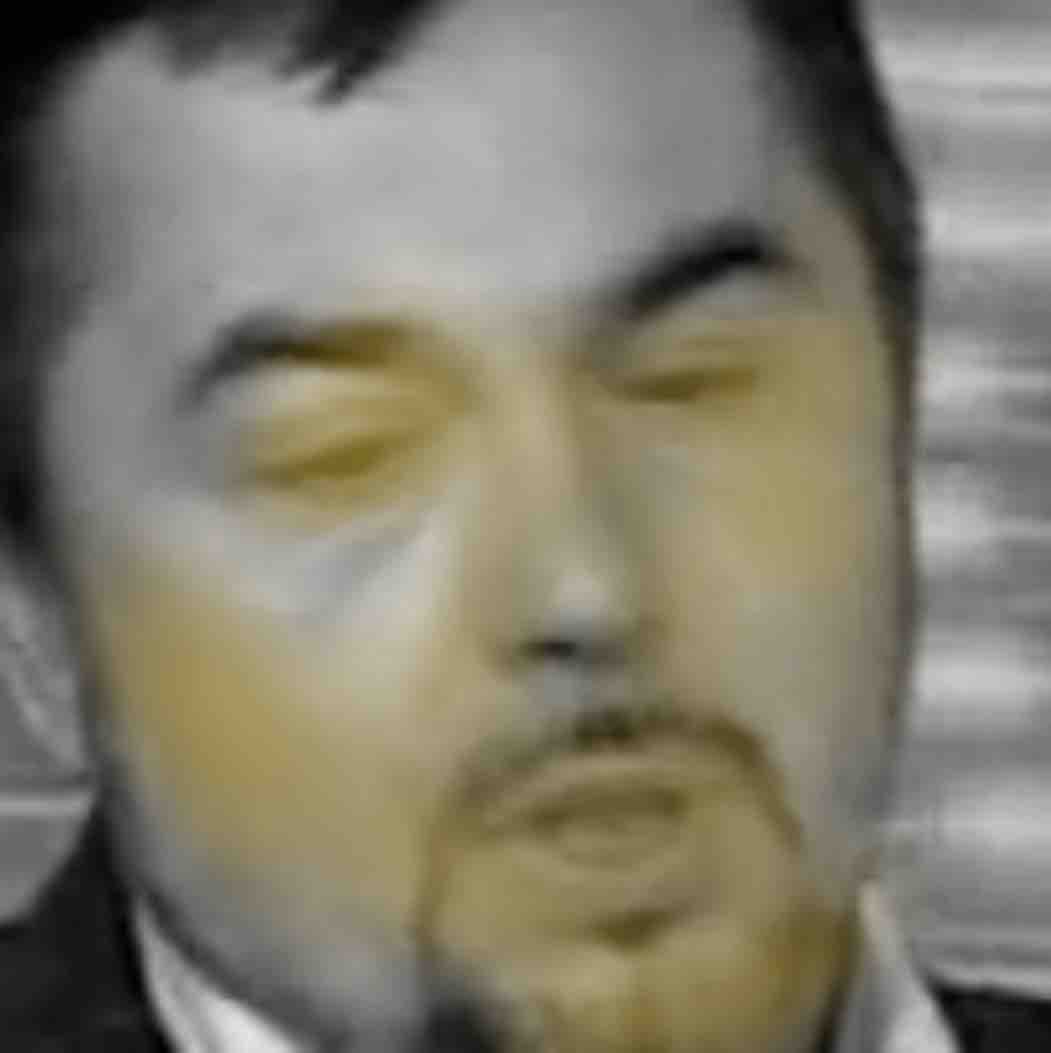}}
 \subfloat[][]{\includegraphics[width=.23\linewidth]{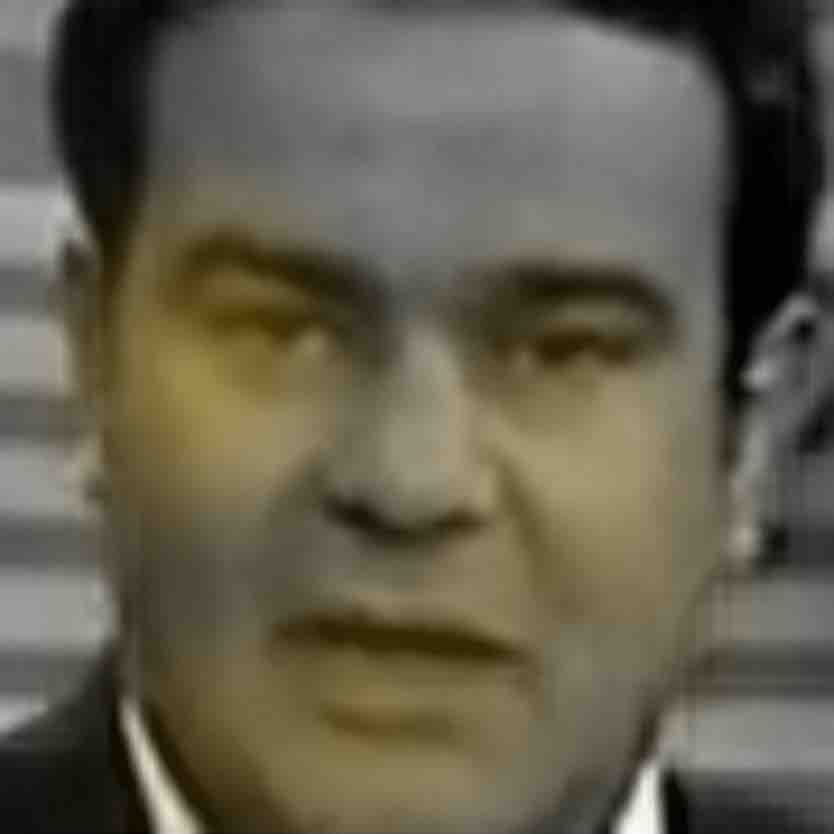}}
   \subfloat[][]{\includegraphics[width=.23\linewidth]{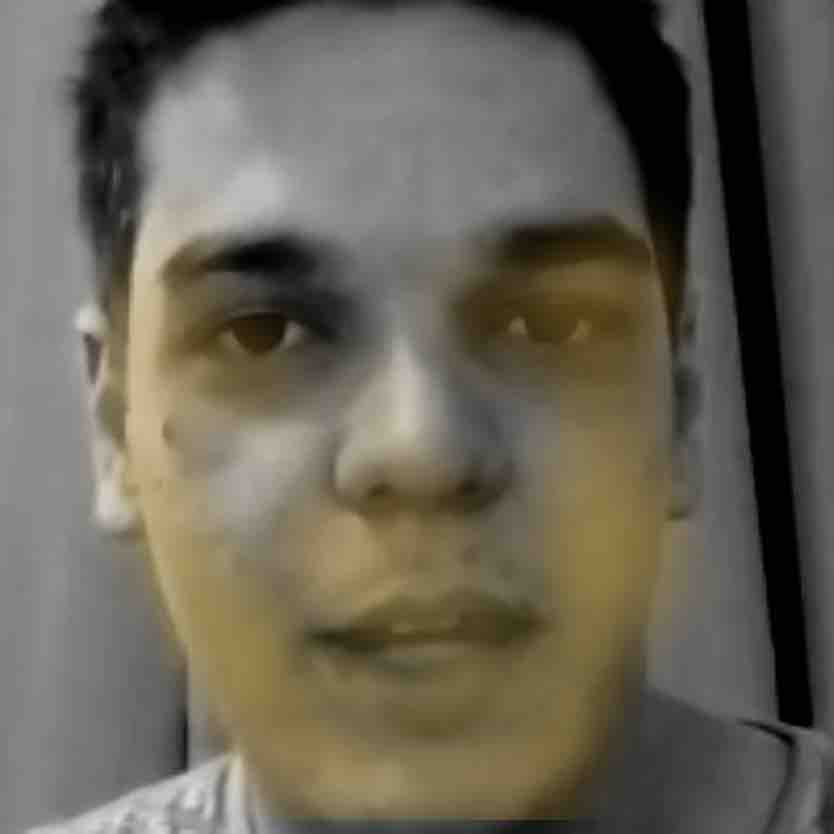}}
 \subfloat[][]{\includegraphics[width=.23\linewidth]{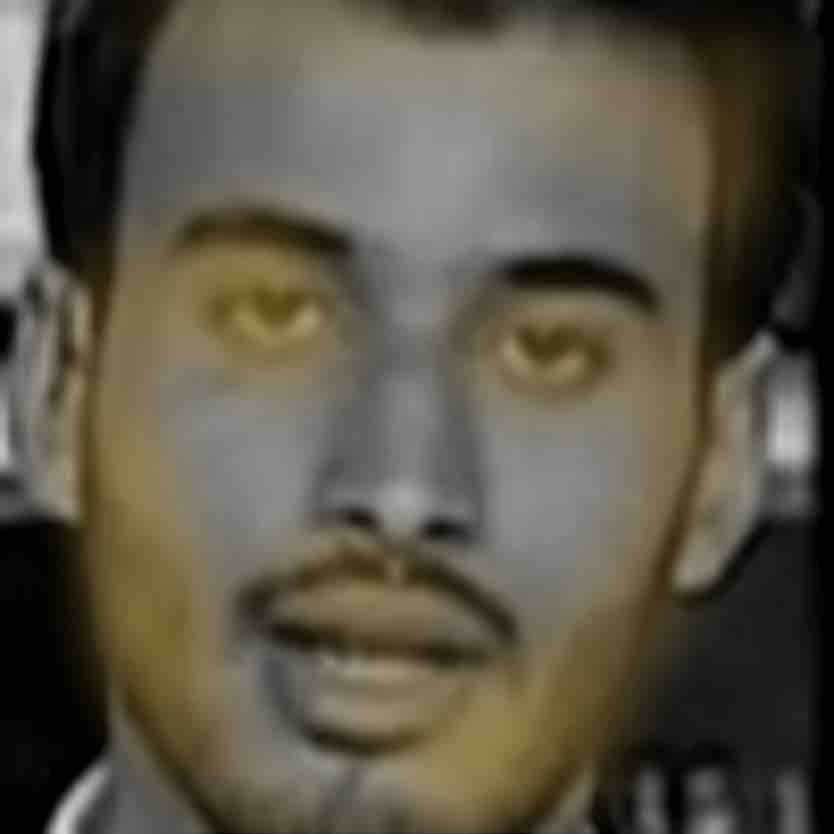}}
\caption{Patch level attention, $\alpha$, (shown in yellow) interpolated in 2D for fake (a)-(d) and real (e)-(h) face images. Brighter intensity values correspond to higher activations. Note that we have displayed images in grayscale for the clarity of the overlaid activations.}
\label{fig:activation_plot_2D}
\end{figure}

In Fig. \ref{fig:activation_plot_2D} we analyse the patch level attention $\alpha$ (i.e Eq. \ref{eq:patch_level_attention}) given to individual face regions of the face for sample images. We populate a 2D heat map using the activations. As $K= 14 \times 14$, we upscale the heat map to fit the original image dimensions. As a result, there is only a rough correspondence. The activation values are shown in yellow and brighter intensity values correspond to higher activations. 

In Fig. \ref{fig:activation_plot_2D} we observe that eye, lip and cheek regions have a high level of attention to measure the image authenticity. These activation plots verify the importance of capturing the facial attributes hierarchically, while preserving their identity and mapping long-term dependencies for accurate detection of fraudulent and tampered faces; and enabling transferability among different types of attacks. 

\section{Relation to Current State-of-the-art}

We draw theoretical comparisons between the proposed HMN module and the state-of-the-art Neural Turing Machine (NTM) \cite{graves2014neural}, Dynamic Memory Networks (DMN) \cite{xiong2016dynamic} and Tree Memory Network (TMN)  \cite{fernando2018tree} methods. 

When comparing with the NTM there exist clear differences between the memory output and memory update procedures. The memory is composed of $l$ slots, let the current state of the $i^{th}$ memory slot at time instance $t-1$ be denoted by $M_i^{t-1}$ and $\gamma$ denotes the normalised attention weight vector, then the memory output of the NTM can be written as,
\begin{equation}
r^t= \sum_{i}\gamma_iM_i^{t-1}.
\end{equation}

The DMN architecture utilised in \cite{xiong2016dynamic} extended this single level attention by manually constructing the element-wise absolute differences and augmenting the memory representation as follows,
\begin{equation}
\mu_i^{t}= [ f_i^{t} \circ q^t; M_i^{t-1} \circ f_i^t; | f_i^{t} - M_i^{t-1} |; | f_i^{t} - q^t | ],
\end{equation}
where $\circ$ denotes an element-wise product and $;$ represents concatenation of the vectors. It should be noted that the authors of \cite{xiong2016dynamic} consider a visual and textual question answering task where there is a seperate question $q_t$ input in addition to the image input which generates $f_i^{t}$. However, in the proposed HMN $f_i^{t}$ formulates $q^t$. Similar to the NTM, the information extracted from the memory module is composed into a single vector, $r^t$, using a single level of attention as,
\begin{equation}
r^t= \sum_{i}\gamma_i\mu_i^{t}.
\end{equation}

Comparing these processes with Eq. \ref{eq:input_level_attention} to \ref{eq:memory_level_attention} it is clear that the proposed method utilises a hierarchical attention structure, ranging from the input level, to the patch level and output level when synthesising the memory output in contrast to the single level attention of NTM and DMN.  

Next we compare the memory update mechanisms between HMN and NTM models. NTM utilises an erase, $e^t$, and add, $a^t$, vectors which contain values ranging between 0 and 1. These are learned by the update controller which learns what potion of the information in a particular memory slot to erase and what portion to add. 

Let $I$ denote a raw vector of 1s, then the two step memory update process of NTM can be written as,
\begin{equation}
\tilde{M_i^t}= M_i^{t-1}[I - \gamma_i^te^t],
\end{equation}
and
\begin{equation}
M_i^t= \tilde{M_i^{t}}+ \gamma_i^ta^t.
\end{equation}

The DMN model in \cite{xiong2016dynamic} updates the memory as,
\begin{equation}
M^t= \mathrm{Relu}(\tilde{W}[M^{t-1}; r^t; q^t]),
\end{equation}
   
where $;$ represents concatenation of the vectors and $\tilde{W}$ is a weight matrix. Once again, these memory update processes are not optimal when working with faces as it can partially update the content in a particular memory slot, which is an embedding of a face, allowing the memory to store unrealistic face attributes. This can be a combination of two faces, which is what is usually occurring with the fake face generation process. Hence, this update procedure contradicts our primary objective of learning natural attributes of real faces. In contrast, the update procedure in Eq. \ref{eq:memory_update} of the proposed method directly concatenates the encoded faces to the memory storage. Hence we preserve the identity of the embeddings. What information should be extracted and when they should be extracted are controlled via the proposed hierarchical operations. 

Furthermore, we would like to draw comparisons with the memory output generation procedure of TMN module. This method utilises structured LSTM (S-LSTM) \cite{zhu2015long} cells arranged in a tree to combine and compress the information stored in memory to an output vector. So the output of the memory, $r^t$, at time instance $t$ is denoted by,

\begin{equation}
r^t=\gamma^tM^{t-1},
\label{eq:tmn}
\end{equation}

where $M^{t-1} \in \mathbb{R}^{k \times 2^{l} -1}$ is the memory matrix that results from concatenating nodes from the tree top to $l= [1, \ldots]$ depth and $k$ is the dimension of of each embedding. 

Even though the hierarchical memory structure demonstrated more robustness than the flat memory architectures such as NTM \cite{graves2014neural} and DMN \cite{xiong2016dynamic}, still this method is not ideal as it mixes embeddings from different faces through the hierarchical feature compression process. These cells combine information from adjacent child nodes and pass a compressed version to the parent node in the tree structured memory, again failing to preserve the identity of the embeddings. In contrast to this single level attention defined in Eq. \ref{eq:tmn}, the proposed method applies hierarchical attention to extract the salient information from faces while preserving the identity of those attributes.

\section{Conclusion} 
Motivated by the social perception and social cognition processes of the human brain, we have presented in this paper a Hierarchical Memory Network (HMN) architecture for the detection of fake and fraudulent faces. The main advantage of this method is the transferability of its learned representation across different, and most importantly, unseen face manipulation methods. By capturing both patch level and image level semantics, and effectively propagating the learned knowledge hierarchically through a memory architecture, the proposed method attains its ability to accurately anticipate the temporal evolution of the observed face, allowing it to discriminate fake faces from real ones. Through extensive evaluations, we have demonstrated the utility of the hierarchical modelling of the stored knowledge while preserving the identity of those facts, and provide visual evidence of how the memory retrieves stored facts while considering how they relate to the current input. This provides solid evidence about the underlying ability of the HMN model to synthesise theories about faces and understand their temporal evolution; abilities which are fundamental across a number of tasks. 


%

%

\ifCLASSOPTIONcompsoc
  \section*{Acknowledgments}
\else
  \section*{Acknowledgment}
\fi

This research was supported by an Australian Research Council (ARC) Discovery grant DP140100793.

\ifCLASSOPTIONcaptionsoff
  \newpage
\fi



\bibliographystyle{IEEEtran}
\bibliography{egbib}

%

\begin{IEEEbiography}[{\includegraphics[width=1in,height=1.25in,clip,keepaspectratio]{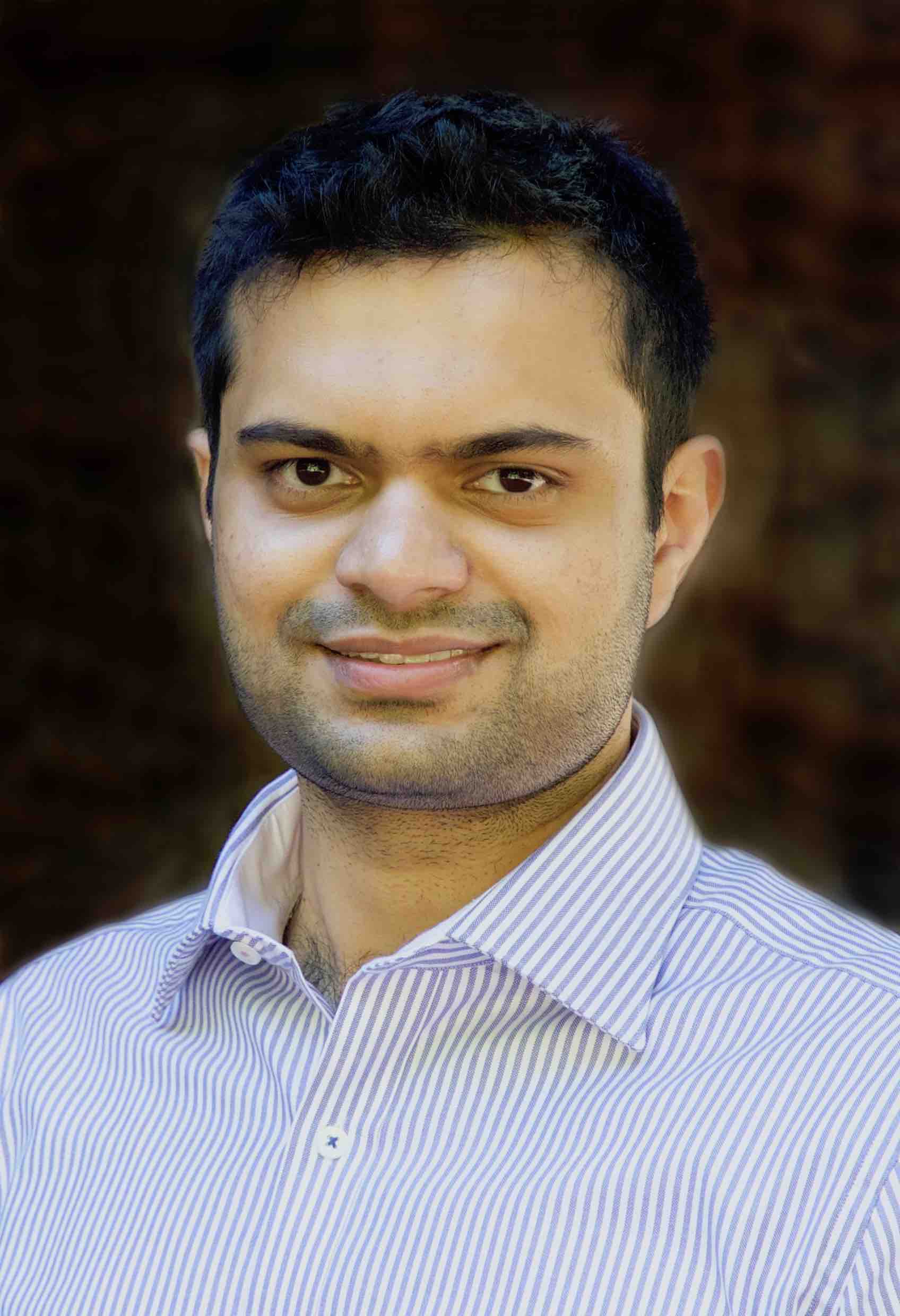}}]{Tharindu Fernando} received his BSc (special degree in computer science) from the University of Peradeniya, Sri Lanka and his PhD from Queensland University of Technology (QUT), Australia, respectively. He is currently a Postdoctoral Research Fellow in the SAIVT Research Program of School Electrical Engineering and Computer Science at QUT. His research interests focus mainly on human behaviour analysis and prediction. 
\end{IEEEbiography}

\begin{IEEEbiography}[{\includegraphics[width=1in,height=1.25in,clip,keepaspectratio]{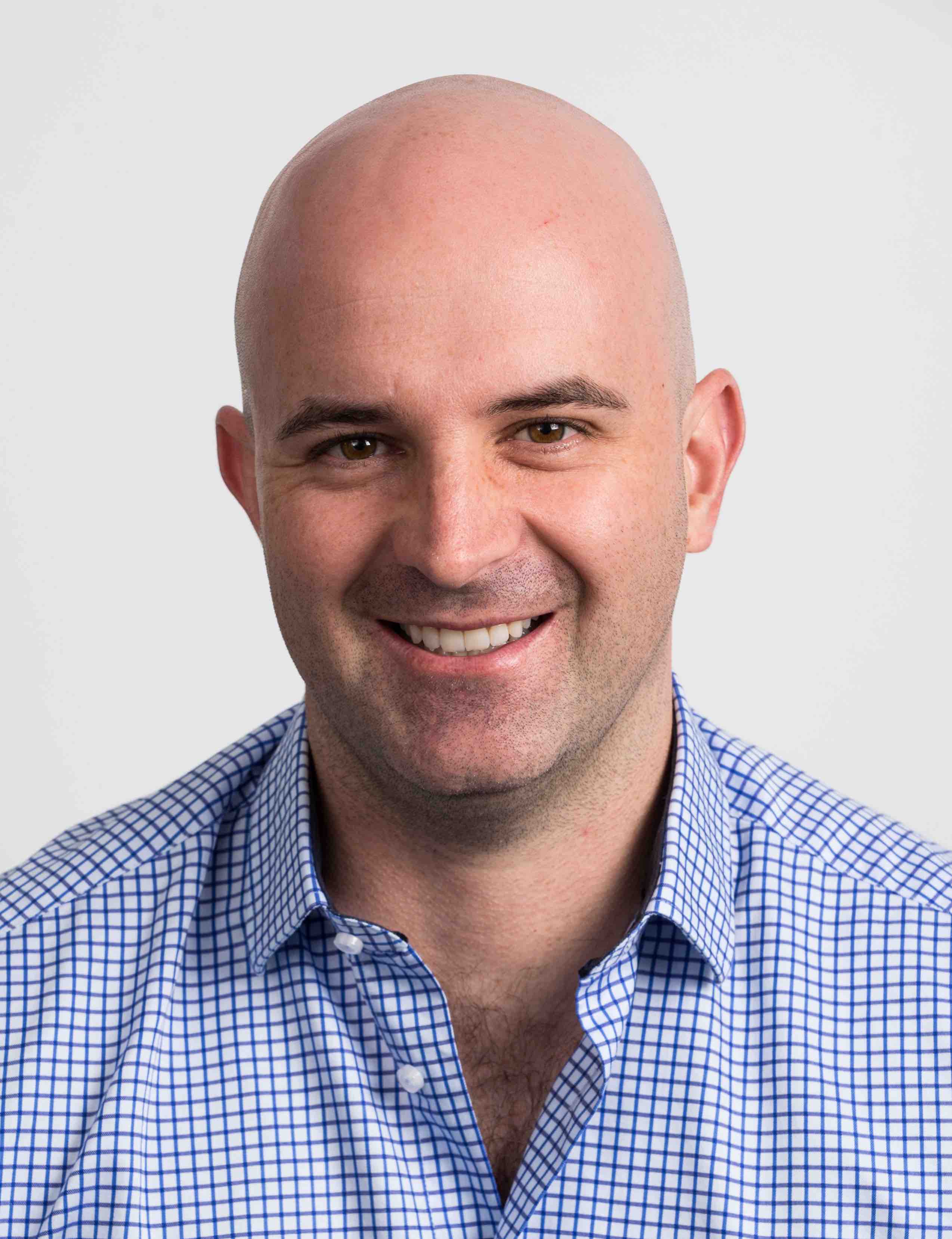}}]{Clinton Fookes} received his B.Eng. (Aerospace/Avionics), MBA, and Ph.D. degrees from the Queensland University of Technology (QUT), Australia. He is currently a Professor and Head of Discipline for Vision and Signal Processing within the Science and Engineering Faculty at QUT. He actively researchers across computer vision, machine learning, and pattern recognition areas. He serves on the editorial board for the IEEE Transactions on Information Forensics \& Security. He is a Senior Member of the IEEE, an Australian Institute of Policy and Science Young Tall Poppy, an Australian Museum Eureka Prize winner, and a Senior Fulbright Scholar.
\end{IEEEbiography}

\begin{IEEEbiography}[{\includegraphics[width=1in,height=1.25in,clip,keepaspectratio]{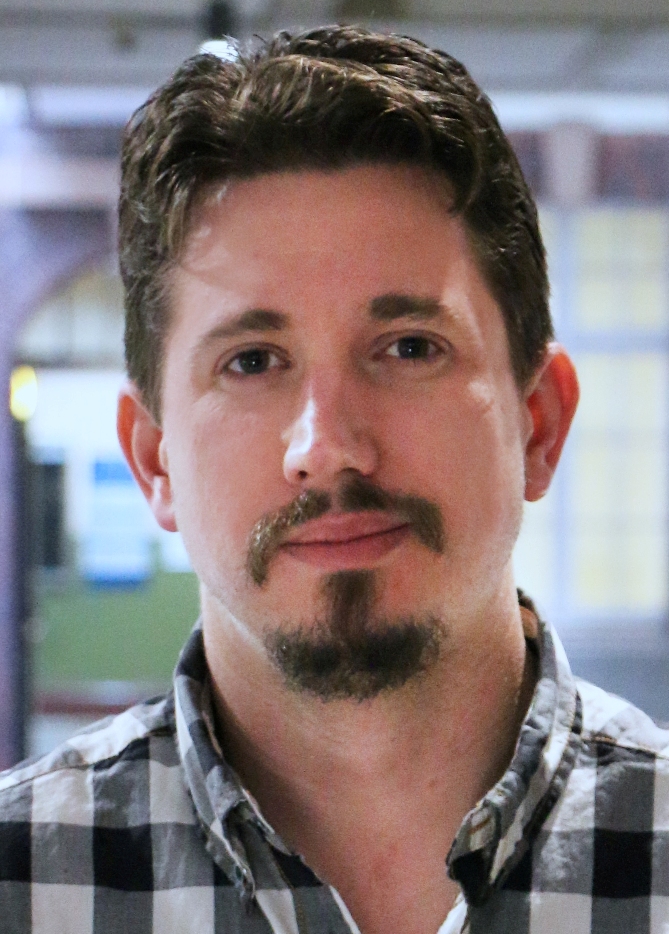}}]{Simon Denman} received a BEng (Electrical), BIT, and PhD in the area of object tracking from the Queensland University of Technology (QUT) in Brisbane, Australia. He is currently a Senior Lecturer within the School of Electrical Engineering and Computer Science at QUT. His active areas of research include intelligent surveillance, video analytics, and video-based recognition.
\end{IEEEbiography}

\begin{IEEEbiography}[{\includegraphics[width=1in,height=1.25in,clip,keepaspectratio]{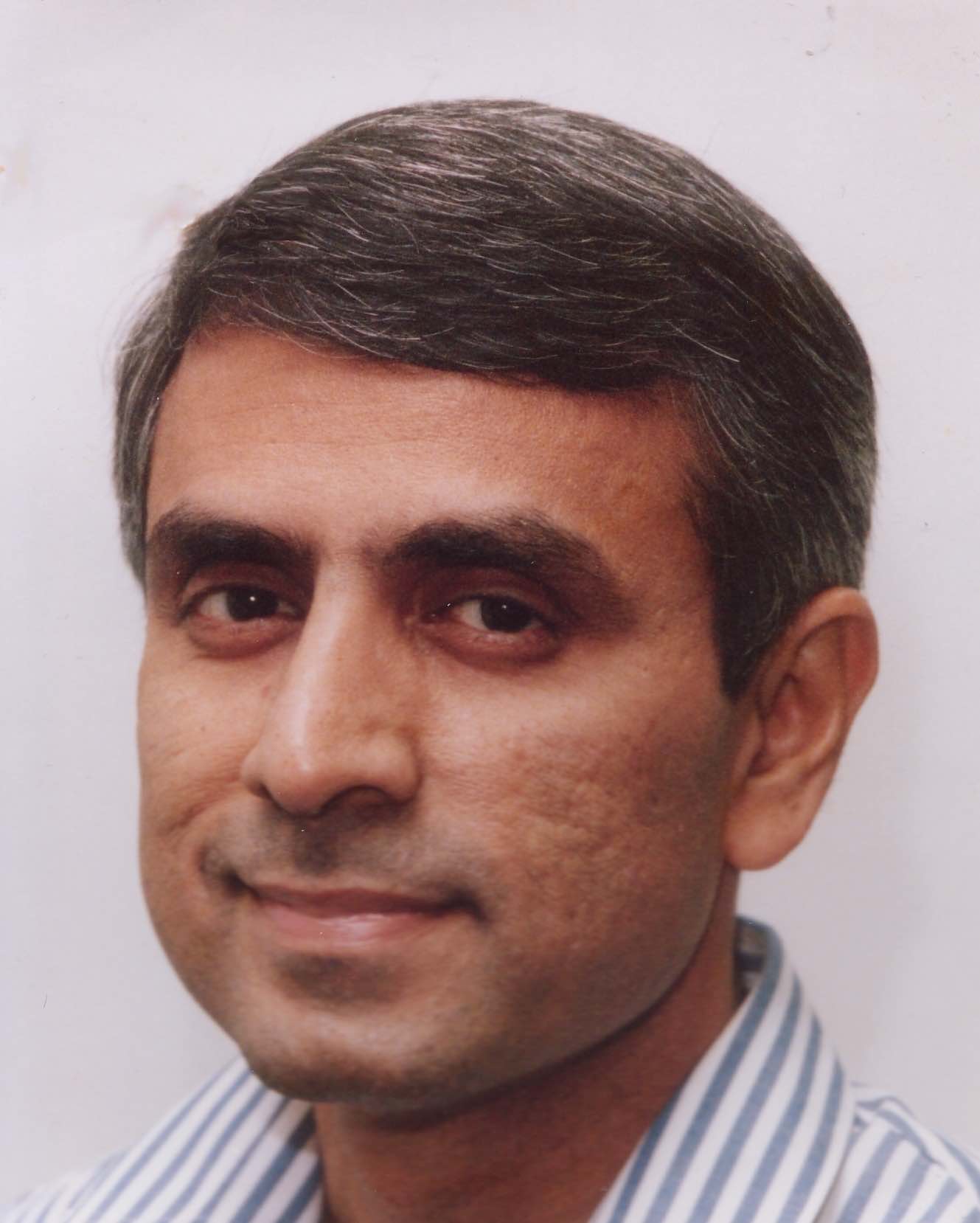}}]{Sridha Sridharan} has a BSc (Electrical Engineering) degree and obtained a MSc (Communication Engineering) degree from the University of Manchester, UK and a PhD degree from University of New South Wales, Australia. He is currently with the Queensland University of Technology (QUT) where he is a Professor in the School Electrical Engineering and Computer Science. Professor Sridharan is the Leader of the Research Program in Speech, Audio, Image and Video Technologies (SAIVT) at QUT, with strong focus in the areas of computer vision, pattern recognition and machine learning. He has published over 600 papers consisting of publications in journals and in refereed international conferences in the areas of Image and Speech technologies during the period 1990-2019. During this period he has also graduated 75 PhD students in the areas of Image and Speech technologies. Prof Sridharan has also received a number of research grants from various funding bodies including Commonwealth competitive funding schemes such as the Australian Research Council (ARC) and the National Security Science and Technology (NSST) unit. Several of his research outcomes have been commercialised.
\end{IEEEbiography}

\end{document}